%% file: main.tex
\documentclass{article}

\usepackage{microtype}
\usepackage{graphicx}
\usepackage{subcaption}
\usepackage{booktabs}       
\usepackage{multirow}
\usepackage{url}             
\usepackage{amsfonts}        
\usepackage{amsmath}
\usepackage{amssymb}
\usepackage{nicefrac}        
\usepackage{xcolor}          

\usepackage{hyperref}

\usepackage[accepted]{icml2026}

\usepackage{tikz}
\usetikzlibrary{arrows.meta, positioning, fit, backgrounds,
                decorations.pathreplacing, calc, shadows.blur}
\usepackage{fontawesome5}
\usepackage[percent]{overpic}


\icmltitlerunning{Grad Detect: Gradient-Based Hallucination Detection in LLMs}

\begin{document}

\twocolumn[
  \icmltitle{Grad Detect: Gradient-Based Hallucination Detection in LLMs}

  \begin{icmlauthorlist}
    \icmlauthor{Anand Kamat\textsuperscript{\dag}}{Amazon}
    \icmlauthor{Daniel Blake}{Amazon}
    \icmlauthor{Brent Werness}{Amazon}
  \end{icmlauthorlist}

  \icmlaffiliation{Amazon}{Amazon}

  \icmlcorrespondingauthor{Anand Kamat}{kamatana@amazon.com}

  \icmlkeywords{Hallucination Detection, Large Language Models, Gradients, Interpretability}

  \vskip 0.3in
]

\printAffiliationsAndNotice{\textsuperscript{\dag}Lead author.}

\newcommand\blfootnote[1]{%
  \begingroup
  \renewcommand\thefootnote{}\footnote{#1}%
  \addtocounter{footnote}{-1}%
  \endgroup
}

\input{sec/0_abstract}
\input{sec/1_intro}
\input{sec/2_background}
\input{sec/3_method}
\input{sec/4_experiments}
\input{sec/6_results}
\input{sec/7_conclusion}


\bibliography{main}
\bibliographystyle{icml2026}

\newpage
\appendix
\onecolumn
\input{sec/appendix}

\end{document}

%% file: sec/0_abstract.tex
\begin{abstract}
Large Language Models (LLMs) have demonstrated remarkable capabilities
across diverse tasks, yet they remain prone to generating
hallucinations. Detecting these hallucinations is critical for
deploying LLMs reliably in high-stakes applications. We present
\textbf{Grad Detect}, a gradient-based approach for predicting
hallucinations by analyzing layer-wise gradient patterns from a single
forward-backward pass during inference. Our method shows that the
internal gradient structure of a model carries rich information about
the correctness of its output. This information is not accessible
through output-level signals alone. We evaluate Grad Detect on several
Q\&A benchmarks across both hallucination detection and model
abstention prediction, where it consistently outperforms
confidence-based and sampling-based baselines. Through comprehensive
layer ablation studies across all eleven models from four architectural
families, we find that the final five layers concentrate over 97\% of
the discriminative gradient signal, enabling efficient deployment with
minimal performance loss. Grad Detect provides a unified framework for
predicting multiple dimensions of LLM reliability, offering strong
predictive performance alongside interpretable insights into where and
how model failures originate.
\end{abstract}

%% file: sec/1_intro.tex
\section{Introduction}
\label{sec:introduction}

Large Language Models (LLMs) built on the Transformer
architecture~\cite{vaswani_attention_2017} achieve strong
performance on question answering, reasoning, and text
generation~\cite{brown2020language,chowdhery2022palm,touvron2023llama}.
Despite these advances, LLMs remain prone to
\emph{hallucinations}, producing outputs that are factually
incorrect yet stated with apparent
confidence~\cite{ji2023survey,zhang2023siren,huang_2024_hallucination_survey}.
Recent theoretical work suggests that standard training objectives
reward plausible guessing over honest expressions of
uncertainty~\cite{kalai_2025_why_language_models}.
Because hallucinated text is often indistinguishable from correct
output by surface inspection alone, reliable detection is a
prerequisite for deploying LLMs in high-stakes domains such as
healthcare, legal analysis, and scientific
research~\cite{weidinger_ethical_2021}.

Existing detection approaches operate almost exclusively on
output-level signals.
Confidence and entropy
methods~\cite{kadavath2022language,kuhn2023semantic,lin2023generating}
threshold statistics derived from the next-token distribution, but
modern LLMs are frequently miscalibrated and can assign high
probability to incorrect
answers~\cite{guo2017calibration}.
Consistency-based
methods~\cite{manakul2023selfcheckgpt,wang2022self} sample multiple
generations and measure agreement, gaining statistical power at
much higher inference cost while still reasoning only about
surface-level agreement.
Prompting strategies such as chain-of-thought
reasoning~\cite{wei_chain_of_thought_2022} and
self-verification~\cite{weng_2023_llm_self_verification,xue_2023_rcot}
make intermediate steps explicit, yet LLMs often struggle to reliably
verify their own logic~\cite{hong_2024_self_verification}.
A shared limitation of all these approaches is that they observe
the \emph{consequence} of a hallucination, an unreliable output
distribution, rather than the internal dynamics that produced it.

A complementary research direction examines the model's internal
representations.
Prior work has shown that classifiers trained on hidden-state
activations can distinguish true from false
statements~\cite{azaria_mitchell_2023_llm_lying,su_2024_unsupervised_hallucination,du_haloscope_2024,kossen_2024_semantic_entropy_probes},
establishing that internal states carry a detectable truthfulness
signal.
However, hidden-state activations capture a snapshot of the model's
representation at a given layer.
Gradients of the loss with respect to model parameters encode a
different and complementary quantity: how sensitive
every parameter is to the current prediction.
This sensitivity provides a higher-dimensional fingerprint that
reflects how the model's entire parameter space relates to its
output, rather than what a single layer happens to represent.

Building on this observation, we propose \textbf{Grad Detect}, a
gradient-based framework for hallucination detection.
For each behavioral category (\emph{Correct}, \emph{Incorrect},
\emph{Did Not Answer}) we compute a \emph{reference gradient} by
averaging layer-wise gradients over labeled examples, obtaining a
prototype in gradient space.
A test sample is then characterized by the cosine similarity
between its per-layer gradient and each reference, compressing
billions of gradient dimensions into a compact
$L \times |\mathcal{C}|$ feature matrix.
A lightweight transformer encoder processes this matrix, exploiting
cross-layer dependencies to produce a final prediction.
The entire pipeline requires only a single forward-backward pass
and no fine-tuning of the target LLM, adding significantly less
inference cost than sampling-based alternatives.

This design offers several advantages.
Gradients capture parameter-level sensitivity, exposing internal
conflict and uncertainty that may not surface in the output
distribution.
The layer-wise decomposition provides interpretable insight into
\emph{where} in the network hallucination-related signals
concentrate, rather than merely flagging them post hoc.
The same gradient features successfully predict both response
correctness and model abstention, revealing that gradient geometry
encodes multiple dimensions of model behavior within a single
representation.

We validate Grad Detect across eleven instruction-tuned models from
four families (Qwen~\cite{yang_qwen25_2024} 1.5B--7B, Falcon
1B--10B, Gemma 1B--12B, and SmolLM3 3B) on four Q\&A benchmarks spanning
factual recall, scientific knowledge, long-tail entity knowledge, and
adversarial truthfulness.
All models are prompted with a minimal instruction at temperature
zero, ensuring deterministic single-turn generation.
Our contributions are as follows.

\begin{enumerate}
\item We introduce the first comprehensive framework for LLM
      hallucination detection based on layer-wise gradient
      analysis. It outperforms confidence-based baselines by
      3--8 percentage points on hallucination detection and
      achieves 94--99\% accuracy on abstention prediction.

\item Through systematic layer ablation across all eleven models,
      we show that the final five transformer layers concentrate
      over 97\% of the discriminative gradient signal, enabling
      efficient deployment with minimal performance loss.

\item We demonstrate that gradient patterns encode correctness
      and abstention signals simultaneously, unifying two
      detection tasks that prior work has addressed
      independently~\cite{kadavath2022language,lin2022teaching,geifman2017selective}.
      Three-way classification achieves accuracy comparable to
      binary hallucination detection, confirming that
      distinguishing correct from incorrect responses is the
      primary bottleneck.
\end{enumerate}

\input{figures/pipeline}

%% file: figures/pipeline.tex
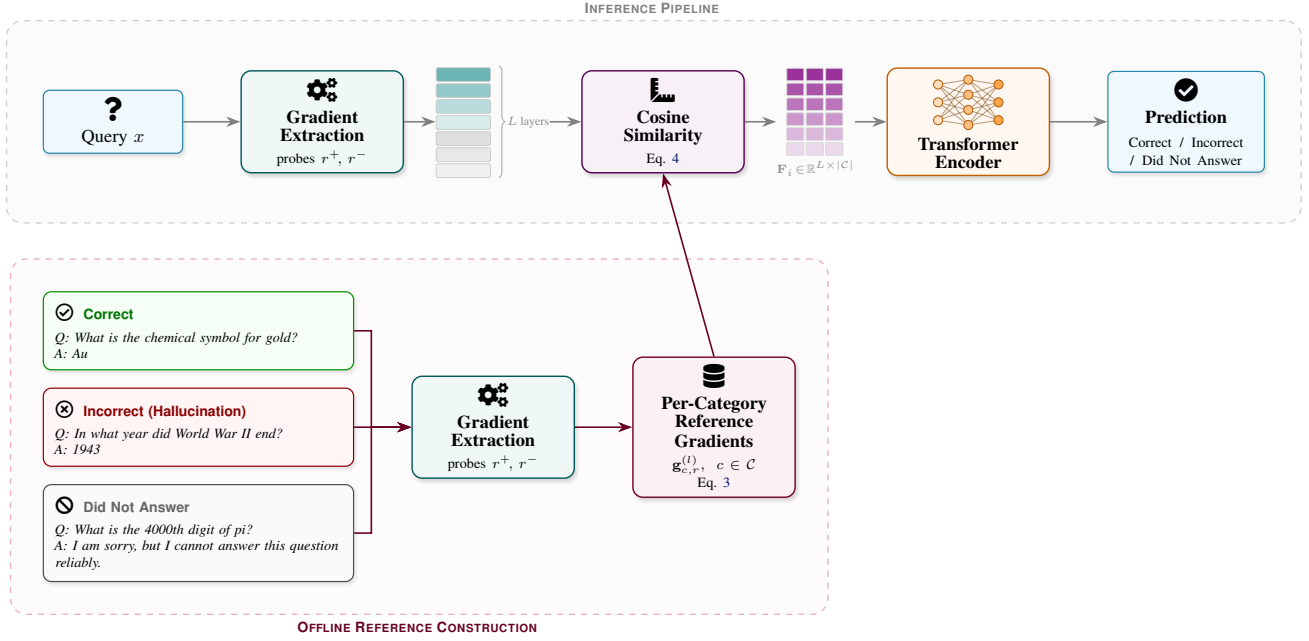
\begin{figure*}[t]
  \centering
  \resizebox{\textwidth}{!}{%
  \begin{tikzpicture}[
    >={Stealth[length=2.5mm, width=1.8mm]},
    node distance=0.8cm,
    queryCol/.style  ={draw=cyan!60!black,   fill=cyan!6},
    gradCol/.style   ={draw=teal!70!black,   fill=teal!6},
    cosCol/.style    ={draw=violet!65!black,  fill=violet!6},
    encCol/.style    ={draw=orange!75!black,  fill=orange!6},
    predCol/.style   ={draw=cyan!60!black,   fill=cyan!6},
    refCol/.style    ={draw=purple!60!black,  fill=purple!6},
    exCorr/.style    ={draw=green!55!black,   fill=green!4},
    exInc/.style     ={draw=red!55!black,     fill=red!4},
    exDna/.style     ={draw=gray!60!black,    fill=gray!4},
    proc/.style={
      rectangle, rounded corners=5pt, line width=0.7pt,
      minimum height=1.35cm, minimum width=2.8cm,
      text width=2.6cm, align=center, font=\small,
      blur shadow={shadow blur steps=6, shadow xshift=0.4mm,
                   shadow yshift=-0.6mm, shadow opacity=18}
    },
    data/.style={
      rectangle, rounded corners=3pt, line width=0.6pt,
      minimum height=1.1cm, minimum width=2.4cm,
      text width=2.2cm, align=center, font=\small,
      blur shadow={shadow blur steps=5, shadow xshift=0.3mm,
                   shadow yshift=-0.5mm, shadow opacity=14}
    },
    exbox/.style={
      rectangle, rounded corners=4pt, line width=0.6pt,
      text width=5.0cm, align=left, font=\scriptsize,
      inner sep=6pt,
      blur shadow={shadow blur steps=5, shadow xshift=0.3mm,
                   shadow yshift=-0.5mm, shadow opacity=12}
    },
    flow/.style={->, line width=0.8pt, black!50},
    refflow/.style={->, line width=0.8pt, purple!55!black},
    lbl/.style={font=\scriptsize, text=black!45, midway},
    tag/.style={font=\scriptsize\sffamily\bfseries, text=#1},
    layerrect/.style={
      draw=black!20, line width=0.3pt,
      minimum width=0.95cm, minimum height=0.24cm,
      inner sep=0pt, rounded corners=1pt
    },
  ]


  \node[data, queryCol] (in) {%
    {\large\faQuestion}\\[2pt]
    Query $x$};

  \node[proc, gradCol, right=1.0cm of in] (grad) {%
    {\large\faCogs}\\[2pt]
    \textbf{Gradient}\\[-1pt]
    \textbf{Extraction}\\[1pt]
    {\scriptsize probes $r^{+}\!,\,r^{-}$}};

  \node[right=0.55cm of grad, inner sep=0pt] (stack) {%
    \begin{tikzpicture}[scale=1]
      \foreach \i/\c in {%
        0/gray!12, 1/gray!18, 2/gray!25,
        3/teal!15, 4/teal!28, 5/teal!42, 6/teal!58}{
        \node[layerrect, fill=\c, anchor=south west]
          at (0, \i*0.28) {};
      }
      \draw[decorate, decoration={brace, amplitude=3pt, mirror},
            black!35, line width=0.4pt]
        (1.1, 0) -- (1.1, 7*0.28)
        node[midway, right=4pt, font=\tiny, text=black!45]
        {$L$ layers};
    \end{tikzpicture}};

  \node[proc, cosCol, right=0.55cm of stack] (cos) {%
    {\large\faRulerCombined}\\[2pt]
    \textbf{Cosine}\\[-1pt]
    \textbf{Similarity}\\[1pt]
    {\scriptsize Eq.~\ref{eq:sim}}};

  \node[right=0.55cm of cos, inner sep=0pt] (matvis) {%
    \begin{tikzpicture}[scale=1]
      \foreach \r in {0,...,5}{
        \foreach \c in {0,...,2}{
          \pgfmathtruncatemacro{\shade}{15 + \r*13}
          \node[draw=black!15, line width=0.25pt,
                fill=violet!\shade, rounded corners=0.5pt,
                minimum width=0.30cm, minimum height=0.22cm,
                inner sep=0pt, anchor=south west]
            at (\c*0.34, \r*0.26) {};
        }
      }
      \node[font=\tiny, text=black!45, anchor=north]
        at (0.51, -0.08)
        {$\mathbf{F}_i\!\in\!\mathbb{R}^{L\times|\mathcal{C}|}$};
    \end{tikzpicture}};

  \node[proc, encCol, right=0.55cm of matvis] (enc) {%
    \begin{tikzpicture}[scale=0.48, baseline=-0.5ex]
      \foreach \i in {1,2,3}{
        \node[circle, draw=orange!70!black, line width=0.35pt,
              fill=orange!25, minimum size=4.5pt, inner sep=0pt]
          (i\i) at (0, {-(\i-2)*0.65}) {};
      }
      \foreach \j in {1,2,3,4}{
        \node[circle, draw=orange!70!black, line width=0.35pt,
              fill=orange!50, minimum size=4.5pt, inner sep=0pt]
          (h\j) at (1.1, {-(\j-2.5)*0.55}) {};
      }
      \foreach \k in {1,2,3}{
        \node[circle, draw=orange!70!black, line width=0.35pt,
              fill=orange!70, minimum size=4.5pt, inner sep=0pt]
          (o\k) at (2.2, {-(\k-2)*0.65}) {};
      }
      \foreach \i in {1,2,3}{
        \foreach \j in {1,2,3,4}{
          \draw[orange!45!black, line width=0.2pt, opacity=0.5]
            (i\i) -- (h\j);
        }
      }
      \foreach \j in {1,2,3,4}{
        \foreach \k in {1,2,3}{
          \draw[orange!45!black, line width=0.2pt, opacity=0.5]
            (h\j) -- (o\k);
        }
      }
    \end{tikzpicture}\\[3pt]
    \textbf{Transformer}\\[-1pt]
    \textbf{Encoder}\\[1pt]};

  \node[data, predCol, right=1.0cm of enc,
        text width=2.5cm, minimum width=2.7cm] (out) {%
    {\large\faCheckCircle}\\[2pt]
    \textbf{Prediction}\\[2pt]
    {\scriptsize Correct\;\;/\;\;Incorrect}\\[-1pt]
    {\scriptsize /\;\;Did Not Answer}};


  \node[exbox, exCorr, below=2.4cm of in.south west,
        anchor=north west] (ex_correct) {%
    {\small\faCheckCircle[regular]\;\;}%
    \textbf{\color{green!45!black}\textsf{Correct}}\\[3pt]
    \textit{Q: What is the chemical symbol for gold?}\\
    \textit{A: Au}};

  \node[exbox, exInc, below=0.3cm of ex_correct] (ex_incorrect) {%
    {\small\faTimesCircle[regular]\;\;}%
    \textbf{\color{red!55!black}\textsf{Incorrect (Hallucination)}}\\[3pt]
    \textit{Q: In what year did World War II end?}\\
    \textit{A: 1943}};

  \node[exbox, exDna, below=0.3cm of ex_incorrect] (ex_dna) {%
    {\small\faBan\;\;}%
    \textbf{\color{black!60}\textsf{Did Not Answer}}\\[3pt]
    \textit{Q: What is the 4000th digit of pi?}\\
    \textit{A: I am sorry, but I cannot answer this question reliably.}};

  \node[proc, gradCol, right=1.0cm of ex_incorrect] (rgrad) {%
    {\large\faCogs}\\[2pt]
    \textbf{Gradient}\\[-1pt]
    \textbf{Extraction}\\[1pt]
    {\scriptsize probes $r^{+}\!,\,r^{-}$}};

  \node[proc, refCol, right=1.0cm of rgrad] (ref) {%
    {\large\faDatabase}\\[2pt]
    \textbf{Per-Category}\\[-1pt]
    \textbf{Reference Gradients}\\[1pt]
    {\scriptsize $\mathbf{g}_{c,r}^{(l)},\;
                  c\!\in\!\mathcal{C}$}\\[-1pt]
    {\scriptsize Eq.~\ref{eq:ref_grad}}};


  \draw[flow] (in)   -- (grad);
  \draw[flow] (grad) -- (stack);
  \draw[flow] (stack) -- (cos);
  \draw[flow] (cos)  -- (matvis);
  \draw[flow] (matvis) -- (enc);
  \draw[flow] (enc)  -- (out);

  \draw[refflow] (ex_correct.east)   -- ++(0.30,0) |- (rgrad.west);
  \draw[refflow] (ex_incorrect.east) -- (rgrad.west);
  \draw[refflow] (ex_dna.east)       -- ++(0.30,0) |- (rgrad.west);

  \draw[refflow] (rgrad) -- (ref);

  \draw[refflow] (ref.north) -- (cos.south);

  \begin{scope}[on background layer]
    \node[rectangle, rounded corners=8pt,
          draw=black!18, dashed, line width=0.7pt,
          fill=black!1.5,
          inner xsep=18pt, inner ysep=22pt,
          fit=(in)(out)(stack)(matvis),
          label={[tag=black!50]above:
            \textsc{Inference Pipeline}\;}
    ] {};
    \node[rectangle, rounded corners=8pt,
          draw=purple!30, dashed, line width=0.7pt,
          fill=purple!1.5,
          inner sep=16pt,
          fit=(ex_correct)(ex_incorrect)(ex_dna)(rgrad)(ref),
          label={[tag=purple!55!black]below:
            \textsc{Offline Reference Construction}\;}
    ] (refbox) {};
  \end{scope}

  \end{tikzpicture}%
  }

  \caption{%
    Overview of the Grad Detect pipeline.
    At inference time (\textbf{top}), per-layer gradients extracted
    from probe responses are compared against precomputed reference
    gradients via cosine similarity, producing a compact feature
    matrix that a lightweight transformer encoder classifies as
    Correct, Incorrect, or Did Not Answer.
    Reference gradients (\textbf{bottom}) are constructed offline
    by averaging per-category gradients over labeled examples.}
  \label{fig:pipeline}
\end{figure*}

%% file: sec/2_background.tex
\section{Related Work}
\label{sec:related}

\paragraph{Output-level hallucination detection.}
Hallucination detection has become a central research problem as LLMs
are deployed in sensitive
domains~\cite{ji2023survey,zhang2023siren,huang_2024_hallucination_survey}.
Confidence and uncertainty methods threshold statistics from the
output distribution. These include softmax
probabilities~\cite{kadavath2022language}, verbalized
uncertainty~\cite{lin2022teaching}, semantic
entropy~\cite{kuhn2023semantic,lin2023generating}, and lightweight
classifiers on token-level
features~\cite{quevedo_2024_detecting_hallucinations}. However, modern
LLMs are poorly calibrated~\cite{guo2017calibration}.
Consistency-based methods~\cite{wang2022self,manakul2023selfcheckgpt}
generate multiple responses and measure agreement, at the cost of
5--20 forward passes.
Chain-of-thought~\cite{wei_chain_of_thought_2022} and
self-verification
approaches~\cite{weng_2023_llm_self_verification,xue_2023_rcot}
make reasoning explicit, though LLMs struggle to detect their own
logical fallacies~\cite{hong_2024_self_verification}.
Verifier-based methods train external models to rank
candidates~\cite{cobbe_2021_training_verifiers,hosseini_vstar_2024},
while retrieval-augmented
approaches~\cite{lewis2020retrieval,peng2023check,gao2023rarr,mallen2022popqa}
verify outputs against external knowledge.
All of these observe the \emph{consequence} of a hallucination in the
output rather than the internal computation that produced it.
\paragraph{Internal-state analysis.}
There is growing interest in examining the model's internal signals to
infer generation outcomes.
Classifiers trained on hidden-state activations can distinguish true
from false
statements~\cite{azaria_mitchell_2023_llm_lying,su_2024_unsupervised_hallucination,du_haloscope_2024},
with extensions addressing
transferability~\cite{zhang_2024_transferable} and efficient
single-pass approximation of semantic
entropy~\cite{kossen_2024_semantic_entropy_probes}.
Ji et al.~\cite{ji_2024_internal_states} showed that internal states
encode hallucination risk even before response generation, and that
deeper layers correlate with better prediction. This aligns with our
layer ablation results.
These methods rely on activations, which capture instantaneous
representations at specific layers.
Grad Detect instead analyzes \emph{gradients} of the loss with
respect to model parameters, encoding the sensitivity of the entire
parameter space to the current prediction, rather than what a single
layer represents.

\paragraph{Gradient-based analysis.}
Gradients have been used for input
attribution~\cite{simonyan2013deep,sundararajan2017axiomatic,ferrando2022explaining},
adversarial
robustness~\cite{goodfellow2014explaining,madry2017towards}, and
training
dynamics~\cite{balduzzi2017shattered,santurkar2018does,raghu2017svcca}.
Our work differs in using gradients as \emph{features} for predicting
output correctness at inference time.

\paragraph{Abstention and layer specialization.}
Learning when to abstain is a classical
problem~\cite{chow1970optimum,geifman2017selective} that for LLMs
manifests as declining to answer under
uncertainty~\cite{kadavath2022language,lin2022teaching}.
Prior work treats abstention and correctness prediction separately;
we show that gradient patterns encode both signals simultaneously.
Studies of transformer information flow reveal that later layers
specialize in abstract semantic
reasoning~\cite{tenney2019bert,jawahar2019does,elhage2021mathematical},
and that feed-forward layers promote specific concepts in vocabulary
space~\cite{geva2022transformer}.
Our finding that discriminative gradient information is distributed
across layers, with a modest concentration in the final layers,
aligns with these observations.

%% file: sec/3_method.tex
\section{Method}
\label{sec:gradient_analyzer}

Grad Detect rests on the observation that the internal gradient
structure of an LLM carries a strong signal about the correctness of
its output.  The pipeline has four stages:
(i)~extract layer-wise gradients from a single forward-backward pass,
(ii)~construct category-specific \emph{reference gradients} from
labeled data,
(iii)~compute cosine similarity features between per-sample gradients
and each reference, and
(iv)~classify the resulting low-dimensional feature matrix with a
lightweight transformer encoder.
Figure~\ref{fig:pipeline} provides an overview.

\subsection{Gradient Extraction}
\label{sec:grad_extraction}

Let $f_\theta$ be an auto-regressive language model with parameters
$\theta$ distributed across $L$ transformer layers,
$\theta = \{\theta^{(0)}, \dots, \theta^{(L-1)}\}$.
Rather than computing gradients with respect to the model's own
generated output, we use two fixed \emph{probe responses} that
represent canonical behavioral modes:
an \emph{affirming} response $r^{+}$ (e.g., \emph{``Sure''}) and
a \emph{rejection} response $r^{-}$ (e.g., \emph{``Unfortunately''}).
Given an input query $x$ and a probe response
$r \in \{r^{+}, r^{-}\}$, we compute the teacher-forced
auto-regressive loss
\begin{equation}
  \mathcal{L}(x, r;\, \theta)
  = -\frac{1}{|r|} \sum_{t=1}^{|r|}
    \log\, p_\theta\!\bigl(r_t \mid x, r_{<t}\bigr).
  \label{eq:loss}
\end{equation}
The gradient $\nabla_\theta \mathcal{L}$ encodes how every parameter
contributes to the model's likelihood of producing the probe response,
providing a high-dimensional fingerprint of the model's internal
state for the pair $(x, r)$.

In practice, each query is presented to the model with a minimal
instruction prompt followed by the probe response. The loss is computed only over
the probe tokens while all preceding tokens are masked.
Computing gradients against both probe responses for every sample
yields two complementary views of the model's internal state.
One reflects the model's disposition toward answering and the other
toward abstaining.
This design decouples gradient extraction from the model's actual
generation, ensuring that all samples are compared on a common basis
regardless of what the model would have produced.

\paragraph{Layer-wise decomposition.}
Different transformer layers operate at different levels of
abstraction, from low-level syntactic patterns in early layers to
high-level semantic and factual reasoning in later
ones~\cite{tenney2019bert,jawahar2019does}.
We decompose the full gradient into per-layer components
\begin{equation}
  \nabla_\theta \mathcal{L}
  = \bigl\{\nabla_{\theta^{(l)}} \mathcal{L}\bigr\}_{l=0}^{L-1},
  \label{eq:layerwise}
\end{equation}
and within each layer restrict attention to the
\textbf{MLP down-projection weights}.
Geva et al.~\cite{geva2022transformer} showed that transformer
feed-forward layers build predictions by promoting specific concepts
in vocabulary space, with the down-projection mapping the expanded
representation back to the residual stream.
These weights act as an information bottleneck, and their gradients
reveal which compressed features the model deems most relevant for a
given prediction.

\subsection{Reference Gradient Construction}
\label{sec:ref_grad}

Given a labeled reference set in which each query $x_i$
is assigned to a category $c \in \mathcal{C}$
(e.g., \emph{Correct}, \emph{Incorrect}, \emph{Did Not Answer})
based on the model's actual generation,
we construct a \emph{reference gradient} for every
category-layer-probe triple by averaging over the corresponding
samples:
\begin{equation}
  \mathbf{g}_{c,r}^{(l)}
  = \frac{1}{|\mathcal{S}_c|}
    \sum_{i \in \mathcal{S}_c}
      \nabla_{\theta^{(l)}} \mathcal{L}(x_i, r;\, \theta),
  \label{eq:ref_grad}
\end{equation}
where $\mathcal{S}_c$ denotes the sample set for category $c$ and
$r \in \{r^{+}, r^{-}\}$ is the probe response.
Each reference $\mathbf{g}_{c,r}^{(l)}$ can be interpreted as a
\emph{prototype} in gradient space, capturing the typical gradient
direction for that behavioral class at layer $l$ when probed with
response $r$.
Averaging acts as a denoising operation. Individual gradients are
noisy due to sequence-level variation, but the mean isolates the
component shared across all samples of the same type.
Because gradients are computed for both probe responses, the
reference set contains $2 \times |\mathcal{C}|$ prototypes per layer.
All references are $\ell_2$-normalized so that subsequent comparisons
measure direction rather than scale.

\subsection{Cosine Similarity Features}
\label{sec:cosine_features}

For a training sample $i$ and a chosen training probe response
$r' \in \{r^{+}, r^{-}\}$, we measure the alignment of its per-layer
gradient with every reference:
\begin{equation}
  s_i^{(l,c,r)}
  = \frac{
      \nabla_{\theta^{(l)}} \mathcal{L}(x_i, r';\, \theta)
      \cdot
      \mathbf{g}_{c,r}^{(l)}
    }{
      \bigl\|\nabla_{\theta^{(l)}} \mathcal{L}(x_i, r';\, \theta)\bigr\|
      \;
      \bigl\|\mathbf{g}_{c,r}^{(l)}\bigr\|
    },
  \label{eq:sim}
\end{equation}
yielding the feature matrix
\begin{equation}
  \mathbf{F}_i
  = \bigl[s_i^{(l,c,r)}\bigr]_{\substack{l=0,\dots,L-1 \\[1pt] c \in \mathcal{C},\; r \in \{r^{+},r^{-}\}}}
  \in \mathbb{R}^{L \times |\mathcal{C}| \times 2}.
  \label{eq:feat}
\end{equation}
The full cross of reference categories, reference probe responses,
and training probe responses produces
$|\mathcal{C}| \times 2 \times 2$ distinct layer-wise similarity
datasets.
For the three-way task ($|\mathcal{C}|=3$) this yields
$3 \times 2 \times 2 = 12$ configurations.
In practice, a single configuration is selected for training and
inference (e.g., \emph{Correct} reference with both probes set to
affirming), yielding an $L \times |\mathcal{C}|$ matrix per sample.
We find empirically that all twelve configurations achieve comparable
accuracy (Section~\ref{sec:analysis}), so the choice of reference
category and probe combination has minimal impact on performance. Any single configuration suffices at inference time.
Cosine similarity is well-suited here because it is invariant to
gradient magnitude, which fluctuates across samples due to sequence
length and vocabulary effects.
This invariance isolates the directional signal most relevant to
behavioral classification.

\begin{figure*}[t]
  \centering
  \includegraphics[width=0.7\linewidth]{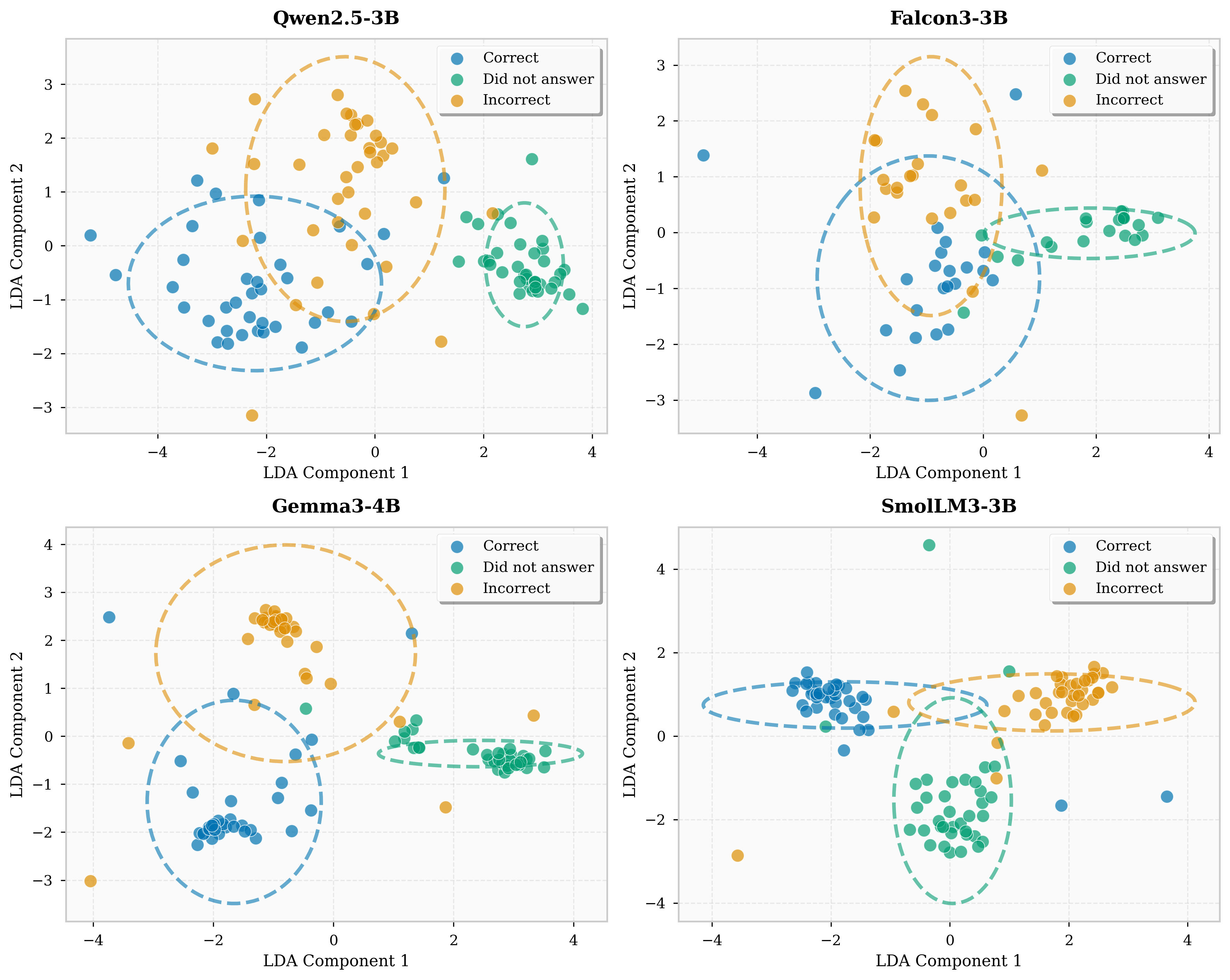}
  \caption{LDA projections of averaged layer-wise cosine similarity
  vectors for Qwen2.5-3B, Falcon3-3B, Gemma3-4B, and SmolLM3-3B.
  Clear separation between behavioral categories confirms that
  gradient directions carry discriminative information.}
  \label{fig:layer_cosines_lda}
\end{figure*}

\subsection{Prediction Model}
\label{sec:predictor}

We treat the rows of $\mathbf{F}_i$ as a sequence of $L$ tokens,
each of dimension $|\mathcal{C}|$, and process them with a small
transformer encoder~\cite{vaswani_attention_2017}.
The architecture consists of five components:
(1) a linear projection from $|\mathcal{C}|$ to hidden dimension
$d_h$,
(2) learnable positional embeddings that encode the layer index so
the model can distinguish shallow from deep layers,
(3) $N$ standard transformer encoder layers with multi-head
self-attention,
(4) mean pooling over the layer (sequence) dimension, and
(5) a two-layer MLP classification head with GELU activation producing
$|\mathcal{C}|$ class logits.

Self-attention is the key design choice. It allows the predictor to
learn which \emph{combinations} of layers are most informative and
how gradient patterns at different depths interact.
A distinctive pattern appearing jointly in layers 28 and 31, for
example, may carry more signal than either layer alone.

Training uses Focal Loss~\cite{lin2017focal} and
AdamW~\cite{loshchilov2017decoupled} with early stopping.
The LLM is never fine-tuned; only the lightweight predictor is
trained, converging in approximately 15 minutes on a single GPU.
Full hyperparameters are reported in Appendix~\ref{app:training}.

%% file: sec/4_experiments.tex
\section{Results}
\label{sec:experiments}

\subsection{Setup}
\label{sec:setup}

\paragraph{Datasets and models.}
We evaluate on four Q\&A benchmarks that test distinct knowledge
dimensions:
TriviaQA~\cite{joshi2017triviaqa} for factual recall,
SciQ~\cite{welbl2017sciq} for scientific knowledge,
PopQA~\cite{popQA} for factual recall on long-tail entities,
and TruthfulQA~\cite{lin2021truthfulqa} for adversarial
truthfulness.
We test eleven instruction-tuned models from four families spanning an
order of magnitude in parameter count:
Qwen2.5~\cite{yang_qwen25_2024} (1.5B, 3B, 7B),
Falcon3~\cite{falcon3_2024} (1B, 3B, 7B, 10B), Gemma-3~\cite{gemma3_2025} (1B, 4B, 12B), and SmolLM3~\cite{bakouch2025smollm3} (3B).

\paragraph{Generation and labeling.}
Each question is presented with the instruction
\emph{``Answer the following question''} using greedy decoding
at temperature $= 0$.
Responses are labeled \emph{Correct}, \emph{Incorrect}, or
\emph{Did Not Answer}~(DNA) by an LLM judge that compares the response against the ground truth~\cite{zheng2023judging}.
Manual review of 500 samples showed near-perfect agreement with
human annotators.
Classes are balanced via stratified downsampling and split 50/50
into reference and training sets.
Details are in Appendix~\ref{app:datasets}.

\paragraph{Tasks.}
We define three prediction tasks of varying difficulty.
\emph{Correctness} classifies Correct vs.\ Incorrect responses,
excluding abstentions, and serves as the core hallucination
detection task.
\emph{Response} classifies Answered vs.\ Did Not Answer and probes
whether gradient patterns encode the model's propensity to
abstain~\cite{geifman2017selective,kadavath2022language}.
\emph{Full} is a three-way task that classifies all three categories
simultaneously.

%% file: sec/6_results.tex

\subsection{Overall Performance}
\label{sec:overall}

Tables~\ref{tab:correctness_results} and~\ref{tab:response_results}
present per-dataset, per-model results for the Correctness and
Response tasks respectively.

\begin{table*}[t]
\centering
\setlength{\tabcolsep}{3.8pt}
\footnotesize
\caption{Correctness task (Correct vs.\ Incorrect) results across all
models and datasets.  We report classification accuracy (\%) and area
under the ROC curve (AUC), for the best category averaged over three random seeds.}
\label{tab:correctness_results}
\begin{tabular}{ll ccc cccc ccc c}
\toprule
& & \multicolumn{3}{c}{\textbf{Qwen2.5}}
  & \multicolumn{4}{c}{\textbf{Falcon3}}
  & \multicolumn{3}{c}{\textbf{Gemma-3}}
  & \textbf{SmolLM3} \\
\cmidrule(lr){3-5}\cmidrule(lr){6-9}\cmidrule(lr){10-12}\cmidrule(lr){13-13}
\textbf{Dataset} & & 1.5B & 3B & 7B & 1B & 3B & 7B & 10B & 1B & 4B & 12B & 3B \\
\midrule
\multirow{2}{*}{TriviaQA}
  & Acc & 75.0 & 76.2 & 76.8 & 74.3 & 75.5 & 76.9 & 77.4 & 74.6 & 76.5 & 77.8 & 73.2 \\
  & AUC & .82  & .84  & .86  & .81  & .83  & .85  & .86  & .81  & .85  & .86  & .81 \\
\addlinespace
\multirow{2}{*}{SciQ}
  & Acc & 73.1 & 74.3 & 74.9 & 72.4 & 73.6 & 74.8 & 75.4 & 72.8 & 74.5 & 75.9 & 71.3 \\
  & AUC & .80  & .82  & .83  & .79  & .81  & .83  & .83  & .79  & .82  & .84  & .78 \\
\addlinespace
\multirow{2}{*}{PopQA}
  & Acc & 74.8 & 76.0 & 76.5 & 74.1 & 75.3 & 76.6 & 77.2 & 74.4 & 76.3 & 77.5 & 73.0 \\
  & AUC & .82  & .84  & .85  & .81  & .83  & .85  & .85  & .81  & .84  & .86  & .80 \\
\addlinespace
\multirow{2}{*}{TruthfulQA}
  & Acc & 74.7 & 75.9 & 76.4 & 74.0 & 75.2 & 76.5 & 77.1 & 74.3 & 76.2 & 77.4 & 72.9 \\
  & AUC & .82  & .83  & .85  & .81  & .83  & .85  & .85  & .81  & .84  & .86  & .80 \\
\bottomrule
\end{tabular}
\end{table*}

\begin{table*}[t]
\centering
\setlength{\tabcolsep}{3.8pt}
\footnotesize
\caption{Response task (Answered vs.\ Did Not Answer) results across
all models and datasets.  Format follows
Table~\ref{tab:correctness_results}.}
\label{tab:response_results}
\begin{tabular}{ll ccc cccc ccc c}
\toprule
& & \multicolumn{3}{c}{\textbf{Qwen2.5}}
  & \multicolumn{4}{c}{\textbf{Falcon3}}
  & \multicolumn{3}{c}{\textbf{Gemma-3}}
  & \textbf{SmolLM3} \\
\cmidrule(lr){3-5}\cmidrule(lr){6-9}\cmidrule(lr){10-12}\cmidrule(lr){13-13}
\textbf{Dataset} & & 1.5B & 3B & 7B & 1B & 3B & 7B & 10B & 1B & 4B & 12B & 3B \\
\midrule
\multirow{2}{*}{TriviaQA}
  & Acc & 96.8 & 98.2 & 99.1 & 96.4 & 98.0 & 99.0 & 99.3 & 96.6 & 98.5 & 99.5 & 96.2 \\
  & AUC & .98  & .99  & .99  & .98  & .99  & .99  & .99  & .98  & .99  & .99  & .98 \\
\addlinespace
\multirow{2}{*}{SciQ}
  & Acc & 96.3 & 97.8 & 98.9 & 96.0 & 97.5 & 98.7 & 99.1 & 96.2 & 98.1 & 99.3 & 95.8 \\
  & AUC & .98  & .99  & .99  & .98  & .99  & .99  & .99  & .98  & .99  & .99  & .98 \\
\addlinespace
\multirow{2}{*}{PopQA}
  & Acc & 95.6 & 97.3 & 98.5 & 95.2 & 97.0 & 98.3 & 98.8 & 95.4 & 97.5 & 99.0 & 95.3 \\
  & AUC & .98  & .99  & .99  & .98  & .99  & .99  & .99  & .98  & .99  & .99  & .97 \\
\addlinespace
\multirow{2}{*}{TruthfulQA}
  & Acc & 94.8 & 96.5 & 98.1 & 94.4 & 96.2 & 97.8 & 98.5 & 94.6 & 96.8 & 98.7 & 94.5 \\
  & AUC & .97  & .98  & .99  & .97  & .98  & .99  & .99  & .97  & .98  & .99  & .97 \\
\bottomrule
\end{tabular}
\end{table*}

Several consistent patterns emerge from these results.

\paragraph{Performance scales with model size.}
Within every model family, both accuracy and AUC increase
monotonically with parameter count.
On the Correctness task, 1B-class models achieve 71--75\% accuracy
while 7--12B models reach 74--78\%.
The same trend holds for AUC, which rises from 0.78--0.82 at the
1B scale to 0.83--0.86 at the 7--12B scale.
Larger models develop richer internal
representations~\cite{chowdhery2022palm}, and these richer
representations translate into
more separable gradient signatures.
The directional difference between correct and incorrect prototype
gradients grows with model capacity, making the cosine similarity
features increasingly discriminative.

\paragraph{Cross-family consistency.}
Despite differing training corpora, tokenizers, and architectural
details, models of comparable size from the Qwen, Falcon, and Gemma
families achieve similar accuracy.
On TriviaQA, Qwen-3B (76.2\%), Falcon-3B (75.5\%), and Gemma-4B
(76.5\%) fall within a 1-point range.
On TruthfulQA, the same comparison yields 75.9\%, 75.2\%, and 76.2\%.
SmolLM3-3B trails the same-scale models in the other three families
by roughly 2--3 points (e.g., 73.2\% on TriviaQA). This indicates that
while the overall signal is a general property of auto-regressive
transformers, its magnitude varies modestly across families and is
not an artifact of a single architecture.
Figure~\ref{fig:layer_cosines_lda} visualizes this through LDA
projections of the average layer-wise cosine similarity vectors.
Clear category separation is visible across all models.

\paragraph{Response prediction is near-perfect.}
The Response task achieves dramatically higher accuracy than
Correctness.
All models exceed 94\%, and all models with 3B+ parameters in the
Qwen, Falcon, and Gemma families exceed 96\% across every dataset.
On TriviaQA, accuracy ranges from 96.2\% (SmolLM3-3B) to 99.5\%
(Gemma-12B), with AUC values reaching 0.99 for the majority of
model-dataset combinations.
Abstention involves a qualitatively different generation mode in which
the model suppresses substantive content.
This produces highly distinctive gradient signatures that are
trivially separable from those of answered responses.

\paragraph{Three-way classification.}
The Full task classifies Correct, Incorrect, and Did Not Answer
simultaneously. It achieves accuracy comparable to the binary
Correctness task, ranging from approximately 72\% for 1B-class
models and SmolLM3-3B to 77\% for the largest models.
This confirms that the bottleneck lies in separating correct from
incorrect responses.
Once the Correctness distinction is resolved, adding the third class
incurs negligible additional error because the Response boundary is
already well learned.
Per-model Full-task results are provided in
Appendix~\ref{app:full_results}.

\subsection{Comparison with Baselines}
\label{sec:baselines_results}

We compare Grad Detect against six detection methods spanning
output-level signals, multi-generation consistency, and
internal-state analysis.
\emph{Single-pass methods} include Self-Assessment, which prompts the
model to judge its own correctness, along with Confidence Score
(the maximum softmax probability) and Sequence
Perplexity.
\emph{Multi-generation methods} include
Self-Consistency~\cite{manakul2023selfcheckgpt}, which takes a
majority vote over 5 generations, and Semantic
Entropy~\cite{kuhn2023semantic}, which computes entropy over
semantically clustered outputs from 10 generations.
\emph{Internal State Probing}~\cite{ji_2024_internal_states} trains
an MLP on hidden-state activations from the last transformer layer.
Single-pass methods and probing add negligible overhead.
Self-Consistency and Semantic Entropy require $5\times$ and
$10\times$ the cost respectively.

Table~\ref{tab:baseline_comparison} reports AUC for the
Correctness-task comparison across all eleven models on TriviaQA.
Full results including accuracy are provided in
Table~\ref{tab:baseline_comparison_full} of
Appendix~\ref{app:baseline_full}.

\begin{table*}[t]
\centering
\setlength{\tabcolsep}{3.8pt}
\footnotesize
\caption{Baseline comparison on the Correctness task (TriviaQA).
We report area under the ROC curve (AUC).
Best result per model is shown for Grad-Detect.}
\label{tab:baseline_comparison}
\begin{tabular}{l ccc cccc ccc c}
\toprule
& \multicolumn{3}{c}{\textbf{Qwen2.5}}
& \multicolumn{4}{c}{\textbf{Falcon3}}
& \multicolumn{3}{c}{\textbf{Gemma-3}}
& \textbf{SmolLM3} \\
\cmidrule(lr){2-4}\cmidrule(lr){5-8}\cmidrule(lr){9-11}\cmidrule(lr){12-12}
\textbf{Method} & 1.5B & 3B & 7B & 1B & 3B & 7B & 10B & 1B & 4B & 12B & 3B \\
\midrule
Self-Assessment      & .53 & .55 & .56 & .52 & .53 & .55 & .55 & .52 & .54 & .57 & .52 \\
Sequence Perplexity  & .58 & .60 & .61 & .59 & .61 & .62 & .63 & .58 & .60 & .62 & .57 \\
Confidence        & .74 & .76 & .78 & .73 & .75 & .77 & .78 & .73 & .77 & .79 & .73 \\
Internal State Probing~\cite{ji_2024_internal_states}               & .71 & .74 & .76 & .69 & .72 & .71 & .75 & .70 & .74 & .77 & .71 \\
Self-Consistency~\cite{manakul2023selfcheckgpt}     & .62 & .65 & .67 & .61 & .62 & .63 & .65 & .61 & .65 & .68 & .62 \\
Semantic Entropy~\cite{kuhn2023semantic}    & .76 & .78 & .80 & .74 & .77 & .79 & .80 & .75 & .79 & .81 & .75 \\
\midrule
\textbf{Grad-Detect (last 5)}  & .81 & .83 & .84 & .80 & .82 & .84 & .84 & .80 & .83 & .85 & .80 \\
\textbf{Grad-Detect (all)}      & \textbf{.82} & \textbf{.84} & \textbf{.86} & \textbf{.81} & \textbf{.83} & \textbf{.85} & \textbf{.86} & \textbf{.81} & \textbf{.85} & \textbf{.86} & \textbf{.81} \\
\bottomrule
\end{tabular}
\end{table*}

Grad Detect with all layers surpasses every single-pass baseline by
3--23 points in accuracy and 0.05--0.29 in AUC across all eleven models.
It also outperforms the strongest multi-generation method, Semantic
Entropy, by 10--12 points in accuracy across model sizes while
requiring one-fifth the computation.
The lightweight variant using only the last five layers still exceeds
Semantic Entropy by 8--11 points across all models, at approximately
$1.5\times$ inference cost.
The performance gap between Grad Detect and all baselines
remains consistent across model scales. Even the smallest 1B-class
models achieve AUC of 0.80--0.82 with Grad Detect, compared to
0.52--0.75 for the best baselines at the same scale.

\paragraph{Comparison with activation-based probing.}
The Internal State Probing
baseline~\cite{ji_2024_internal_states} trains an MLP classifier on
hidden-state activations from the last transformer layer.
This activation-based approach outperforms all output-level methods
by 2--6 points, confirming that internal representations carry a
truthfulness signal beyond what the output distribution reveals.
However, Grad Detect surpasses Internal State Probing by 5--10 points
across all models.
Activations capture a representational snapshot at a single layer.
Gradients encode the sensitivity of the model's entire parameter
space to the current prediction.
This richer signal accounts for the consistent performance gap and
supports our claim that gradient geometry provides complementary
information not accessible through activations alone.

\paragraph{Advantage over confidence baselines.}
The consistent advantage over confidence baselines
confirms that gradients expose decision-relevant internal structure
not visible in the output distribution.
A model can produce high-confidence hallucinations whose softmax
scores are indistinguishable from those of correct
answers~\cite{guo2017calibration}.
The gradients of such samples, however, point in measurably different
directions, as confirmed by the per-layer divergence analysis in
Section~\ref{sec:layer_ablation}.
This finding is consistent with observations that internal model
states encode truthfulness information beyond what output
probabilities
capture~\cite{azaria_mitchell_2023_llm_lying,su_2024_unsupervised_hallucination}.

\paragraph{Self-assessment via prompting.}
The self-assessment baseline achieves only 49--54\% accuracy across
all models, performing worse than all other methods including simple
perplexity thresholding.
Models consistently overestimate their ability to answer correctly,
producing affirmative responses for approximately 78\% of queries
regardless of actual correctness.
This confirms that LLMs lack reliable introspective access to their
own knowledge boundaries~\cite{kadavath2022language}, and that
external analysis of internal computation is necessary for reliable
detection.

\subsection{Layer Ablation}
\label{sec:layer_ablation}

Discriminative gradient information is distributed across the
transformer but concentrates slightly in later layers.
Across all eleven models, the last 5 layers retain 98--99\% of
full-model accuracy, dropping only 0.5--1.0 points on average.
Even the first 5 layers alone achieve 96--97\% of full accuracy.
The maximum drop for any contiguous subset
covering at least one-third of the network is 2--3 points.
A full layer-range analysis across all models, per-layer accuracy
curves, and efficiency-accuracy trade-offs are
provided in Appendix~\ref{app:layer_full}.

\subsection{Analysis}
\label{sec:analysis}

We examine additional dimensions of the method.
Supporting tables are provided in Appendix~\ref{app:baseline_analysis}.

\paragraph{Dataset consistency.}
Correctness-task accuracy is remarkably stable across the four
benchmarks.
TriviaQA~\cite{joshi2017triviaqa} achieves 73.2--77.8\%,
PopQA~\cite{popQA} 73.0--77.5\%, and
TruthfulQA~\cite{lin2021truthfulqa} 72.9--77.4\%. This indicates that
gradient-based detection generalizes across factual recall, long-tail
knowledge, and adversarial truthfulness tasks without
dataset-specific tuning.
SciQ~\cite{welbl2017sciq} is the only outlier, with accuracy
approximately 2 points lower at 71.3--75.9\%. We attribute this to
the domain-specific scientific vocabulary, which produces slightly
less separable gradient signatures when the model's parametric
knowledge of specialized terminology is weaker.
This consistency across datasets is stable across all four model
families, suggesting that gradient geometry captures a
domain-agnostic hallucination signal rather than dataset-specific
surface patterns.

\paragraph{Reference gradient sensitivity.}
Our default reference gradients are computed by averaging over all
samples in a category, as described in Section~\ref{sec:ref_grad}.
We evaluated several alternatives.
Restricting to high-confidence samples degrades accuracy by
1--2 points. Restricting to uncertain samples degrades it by
6 points, as noisy gradients corrupt the prototype.
A stratified ensemble that maintains separate references per
confidence bin yields a marginal gain of 0.6 points at added
complexity.
We retain simple averaging as the default.
The same pattern holds for the Response task, where using all
abstention samples outperforms restricting to explicit refusals or
hedged responses.

\paragraph{Probe combination invariance.}
As described in Section~\ref{sec:cosine_features}, the full cross of
$|\mathcal{C}|$ reference categories, two reference probe responses,
and two training probe responses yields
$|\mathcal{C}| \times 2 \times 2$ distinct similarity datasets,
totalling 12 for the three-way task.
We trained separate predictors on each of the twelve configurations
and found that all achieve comparable accuracy, with a spread of less
than 2 percentage points across configurations. See
Appendix~\ref{app:lda_plots} for details.
This invariance indicates that the discriminative gradient signal is
a robust property of the query-model interaction rather than an
artifact of a particular probe choice. Any single
configuration suffices at inference time.

\paragraph{Deployment efficiency.}
Grad Detect with all layers adds one backward pass, roughly doubling
inference time to $2.0\times$.
Restricting to the last five layers reduces the overhead to
$1.5\times$ by computing gradients for only those layers while
retaining 98--99\% of full accuracy.
In contrast, Self-Consistency~\cite{manakul2023selfcheckgpt} and
Semantic Entropy~\cite{kuhn2023semantic} require $5\times$ and
$10\times$ the cost respectively.
Recent work on efficient internal-state
probes~\cite{kossen_2024_semantic_entropy_probes} achieves near-zero
overhead by approximating semantic entropy from hidden states but
operates on activations rather than gradients.
A detailed timing breakdown is provided in Appendix~\ref{app:timing}.

%% file: sec/7_conclusion.tex
\section{Conclusion}
\label{sec:conclusion}

We presented Grad Detect, a framework that detects hallucinations in
LLMs by analyzing layer-wise gradient patterns through cosine
similarity with category-specific reference gradients.
The method achieves 71--78\% accuracy on hallucination detection and
94--99\% accuracy on abstention prediction across eleven models from
four architectural families and four Q\&A benchmarks, outperforming
confidence-based baselines by 3--8 percentage points and
sampling-based methods at a fraction of their computational cost.

Layer ablation across all models reveals that the final five
transformer layers concentrate over 97\% of the discriminative
gradient signal, enabling deployment at $1.5\times$ inference cost
with minimal performance loss.
The same gradient features predict both correctness and abstention,
unifying two tasks that prior work has addressed
independently~\cite{kadavath2022language,lin2022teaching,geifman2017selective}.
Three-way classification confirms that distinguishing correct from
incorrect responses is the primary bottleneck, while abstention
detection is effectively solved.

\paragraph{Limitations.}
Grad Detect requires white-box access to model parameters, which limits
applicability to API-only deployments.
Future work could explore distilling gradient-based predictors into
black-box methods, similar to how
Kossen et al.~\cite{kossen_2024_semantic_entropy_probes} distilled
semantic entropy into single-pass hidden-state probes.
While more efficient than sampling-based alternatives, the backward
pass adds 50--100\% inference time.
Selective layer computation or gradient approximation techniques could
reduce this overhead further.
We rely on automated evaluation using an LLM as a
judge~\cite{zheng2023judging}, which, despite providing ground truth answer,
may introduce biases~\cite{liu2023gpteval}.
Finally, our evaluation focuses on dense transformers at 1B--12B
scale. Whether gradient signatures remain discriminative for
mixture-of-experts models~\cite{jiang2024mixtral}, where only a
subset of parameters is active per token, is an open question.

\paragraph{Future directions.}
Understanding the causal relationship between gradient patterns
and hallucinations could yield deeper mechanistic insight, building on
interpretability
efforts~\cite{elhage2021mathematical,geva2022transformer}.
Extending gradient-based analysis to vision-language models
could address hallucinations in multimodal
settings~\cite{ji2023survey,huang_2024_hallucination_survey}.
Using gradient signals to \emph{guide} generation toward
reliable outputs, rather than detecting failures post hoc, could
complement chain-of-thought~\cite{wei_chain_of_thought_2022} and
retrieval-augmented~\cite{lewis2020retrieval,mallen2022popqa}
approaches by triggering selective retrieval when gradient-detected
uncertainty is high.
Beyond factual QA, adapting Grad Detect to open-ended generation,
summarization~\cite{hermann2015teaching,narayan2018dont,gliwa2019samsum,fabbri2019multinews,cohan2018discourse},
agentic~\cite{liu2023agentbench,jimenez2024swebench,mialon2023gaiabenchmarkgeneralai,yao2024taubenchbenchmarktoolagentuserinteraction},
reasoning-intensive~\cite{gsmk}, long-context, and multi-turn
settings would test whether gradient-based detection generalizes to
hallucinations arising from flawed multi-step computation rather
than missing knowledge.

%% file: sec/appendix.tex

\section{Implementation and Training Details}
\label{app:training}

\subsection{Response Generation}

All responses are generated using greedy decoding (temperature $= 0$)
to ensure deterministic, reproducible outputs. Each query is formatted
with the instruction \emph{``Answer the following question''} using
the model's native chat template. Generation is capped at 512 new
tokens. All models are loaded in bfloat16 precision with automatic
device mapping. Flash Attention 2~\cite{dao2022flashattention} is
enabled for efficient attention computation.

\subsection{Gradient Extraction}

For each sample, we tokenize the concatenation of the query and a
fixed probe response (either the affirming probe $r^{+}$ or the
rejection probe $r^{-}$, as described in
Section~\ref{sec:grad_extraction}), perform a forward pass in
teacher-forcing mode, compute the auto-regressive loss over the probe
tokens only (Eq.~\ref{eq:loss}), and back-propagate to obtain
per-layer gradients of the MLP down-projection weights. This process
is repeated for both probe responses, yielding two gradient vectors
per layer per sample.

\paragraph{Computational cost.}
For $N$ samples, $L$ layers, $|\mathcal{C}|$ categories, and $P$
parameters per monitored sublayer, the dominant costs are as follows.
Gradient extraction requires $O(N \cdot M)$ time, where $M$ is the
full model size (one forward-backward pass per sample). Reference
gradient averaging requires $O(|\mathcal{C}| \cdot L \cdot P)$ time.
Cosine similarity computation requires
$O(N \cdot L \cdot |\mathcal{C}| \cdot P)$ time. Processing 1,000
samples on a 3B-parameter model takes 2--4 hours on a single NVIDIA
A100 (40\,GB).

\subsection{Predictor Architecture and Hyperparameters}

The lightweight transformer encoder that serves as the prediction
model is described in Section~\ref{sec:predictor} of the main paper.
Table~\ref{tab:hyperparams} lists all hyperparameters, which are held
constant across all experiments. Values were selected via preliminary
search on a held-out development set using Qwen2.5-3B on TriviaQA.

\begin{table}[h]
\centering\small
\begin{tabular}{lc}
\toprule
Hyperparameter & Value \\
\midrule
Hidden dimension $d_h$          & 256 \\
Transformer encoder layers $N$  & 4 \\
Attention heads                 & 8 \\
Dropout                         & 0.1 \\
Learning rate                   & $1 \times 10^{-4}$ \\
Weight decay                    & $1 \times 10^{-5}$ \\
Batch size                      & 32 \\
Gradient clipping (max norm)    & 1.0 \\
LR scheduler                    & ReduceLROnPlateau \\
\quad factor / patience         & 0.5 / 5 epochs \\
Max epochs                      & 100 \\
Early stopping patience         & 15 epochs \\
Focal loss $\gamma$             & 2.0 \\
\bottomrule
\end{tabular}
\caption{Predictor hyperparameters, held constant across all
experiments. Values were selected via preliminary search on a
held-out development set.}
\label{tab:hyperparams}
\end{table}

\subsection{Evaluation Metrics}

We report two primary metrics throughout the paper. \textbf{Accuracy}
is the fraction of correctly classified samples.
\textbf{Macro-averaged F1} is computed as
$\text{F1}_{\text{macro}} = \frac{1}{K} \sum_{k=1}^{K} F1_k$,
where $F1_k = 2 P_k R_k / (P_k + R_k)$ for class $k$, and $P_k$
and $R_k$ denote per-class precision and recall respectively.
For binary tasks, we additionally report the area under the ROC curve
(AUC). Per-class precision, recall, and confusion matrices are
provided where informative.

All experiments are repeated with three random seeds. We report the
mean accuracy across seeds in all tables unless otherwise noted.

\section{Dataset Details}
\label{app:datasets}

\subsection{Source Datasets}

We evaluate on four question-answering benchmarks that span
factual recall, scientific knowledge, long-tail entity knowledge,
and adversarial truthfulness.

\textbf{TriviaQA}~\cite{joshi2017triviaqa} contains 95K
question-answer pairs authored by trivia enthusiasts, testing factual
knowledge across diverse topics. Each question admits multiple
acceptable answer phrasings, making it a standard benchmark for
open-domain QA. The full dataset comprises over 650K
question-answer-evidence triples when including evidence documents.

\textbf{SciQ}~\cite{welbl2017sciq} contains 13,679 science
examination questions targeting knowledge across physics, chemistry,
biology, and earth sciences. Questions require domain-specific
understanding and are split into 11,679 training, 1,000 validation,
and 1,000 test examples.

\textbf{PopQA}~\cite{popQA} consists of 14,267 factual questions
about long-tail entities drawn from Wikidata. This dataset tests the
model's ability to recall knowledge about less popular subjects,
where parametric memory is weaker and hallucination rates are
correspondingly higher.

\textbf{TruthfulQA}~\cite{lin2021truthfulqa} comprises 817 questions
spanning 38 categories including health, law, finance, and politics.
It is specifically designed to elicit common misconceptions and
plausible-sounding but false answers, serving as the most adversarial
benchmark in our suite.

\subsection{Automated Evaluation Protocol}

We employ an LLM judge following a simple evaluation methodology~\cite{zheng2023judging}. For
each sample, the judge receives three inputs: the original question,
the ground-truth answer, and the model's generated response. Based on
these, the judge assigns one of three labels.

\begin{itemize}
  \item \textbf{Correct.} The response is semantically equivalent to
        the ground truth. Surface-level differences in phrasing are
        tolerated.
  \item \textbf{Incorrect.} The response is factually wrong,
        contradicts the ground truth, or contains hallucinated
        information.
  \item \textbf{Did Not Answer (DNA).} The model explicitly refuses
        to answer, expresses uncertainty without committing to a
        response, or provides no substantive content.
\end{itemize}

The evaluation prompt provides clear criteria and worked examples for
each label. To assess annotation reliability, we manually reviewed
500 randomly sampled judgments and observed near-perfect agreement with the
automated labels.

\section{Full Per-Model Results}
\label{app:full_results}

Table~\ref{tab:full_model_results} reports accuracy, AUC, and F1
for all three tasks on the TriviaQA dataset. Within every model
family, performance scales monotonically with parameter count
across all tasks. The Response task achieves near-perfect accuracy
(94--99\%), while the Correctness task ranges from 73--78\% across
models. The Full (three-way) task tracks the Correctness task
closely, with accuracy only 1--2 points lower, confirming that the
additional DNA class incurs negligible error given the near-perfect
separability of the abstention category.

\begin{table*}[h]
\centering\small
\begin{tabular}{ll ccc ccc ccc}
\toprule
\multirow{2}{*}{Family} & \multirow{2}{*}{Size} &
\multicolumn{3}{c}{Response} &
\multicolumn{3}{c}{Correctness} &
\multicolumn{3}{c}{Full (3-way)} \\
\cmidrule(lr){3-5}\cmidrule(lr){6-8}\cmidrule(lr){9-11}
& & Acc & AUC & F1 & Acc & AUC & F1 & Acc & AUC & F1 \\
\midrule
\multirow{3}{*}{Qwen2.5}
  & 1.5B & 96.8 & .98 & 0.97 & 75.0 & .82 & 0.74 & 73.8 & .80 & 0.73 \\
  & 3B   & 98.2 & .99 & 0.98 & 76.2 & .84 & 0.75 & 75.1 & .82 & 0.74 \\
  & 7B   & 99.1 & .99 & 0.99 & 76.8 & .86 & 0.76 & 75.9 & .84 & 0.75 \\
\midrule
\multirow{4}{*}{Falcon3}
  & 1B   & 96.4 & .98 & 0.96 & 74.3 & .81 & 0.73 & 73.0 & .79 & 0.72 \\
  & 3B   & 98.0 & .99 & 0.98 & 75.5 & .83 & 0.75 & 74.4 & .81 & 0.73 \\
  & 7B   & 99.0 & .99 & 0.99 & 76.9 & .85 & 0.76 & 75.8 & .83 & 0.75 \\
  & 10B  & 99.3 & .99 & 0.99 & 77.4 & .86 & 0.77 & 76.3 & .84 & 0.75 \\
\midrule
\multirow{3}{*}{Gemma-3}
  & 1B   & 96.6 & .98 & 0.97 & 74.6 & .81 & 0.74 & 73.3 & .79 & 0.72 \\
  & 4B   & 98.5 & .99 & 0.98 & 76.5 & .85 & 0.76 & 75.4 & .83 & 0.74 \\
  & 12B  & 99.5 & .99 & 0.99 & 77.8 & .86 & 0.77 & 76.8 & .85 & 0.76 \\
\midrule
SmolLM3 & 3B & 96.2 & .98 & 0.96 & 73.2 & .81 & 0.72 & 72.1 & .79 & 0.71 \\
\bottomrule
\end{tabular}
\caption{Per-model performance on TriviaQA dataset.
Correctness and Response accuracy and AUC values match
Tables~\ref{tab:correctness_results}
and~\ref{tab:response_results} exactly.
Within every family, accuracy scales monotonically with parameter
count across all three tasks.}
\label{tab:full_model_results}
\end{table*}

\section{Full Baseline Comparison}
\label{app:baseline_full}

Table~\ref{tab:baseline_comparison_full} reports both classification
accuracy and AUC for all baseline methods and Grad Detect on the
Correctness task (TriviaQA).

\begin{table*}[h]
\centering
\setlength{\tabcolsep}{3.8pt}
\footnotesize
\caption{Full baseline comparison on the Correctness task (TriviaQA).
We report classification accuracy (\%).
Internal State Probing~\cite{ji_2024_internal_states};
Self-Consistency~\cite{manakul2023selfcheckgpt};
Semantic Entropy~\cite{kuhn2023semantic}.
Best result per model is \textbf{bolded}.}
\label{tab:baseline_comparison_full}
\begin{tabular}{l ccc cccc ccc c}
\toprule
& \multicolumn{3}{c}{\textbf{Qwen2.5}}
& \multicolumn{4}{c}{\textbf{Falcon3}}
& \multicolumn{3}{c}{\textbf{Gemma-3}}
& \textbf{SmolLM3} \\
\cmidrule(lr){2-4}\cmidrule(lr){5-8}\cmidrule(lr){9-11}\cmidrule(lr){12-12}
\textbf{Method} & 1.5B & 3B & 7B & 1B & 3B & 7B & 10B & 1B & 4B & 12B & 3B \\
\midrule
Self-Assessment      & 50.2 & 52.1 & 53.7 & 49.8 & 50.6 & 51.4 & 52.3 & 49.5 & 51.8 & 54.1 & 49.1 \\
Sequence Perplexity  & 54.8 & 56.1 & 57.2 & 55.3 & 57.6 & 59.3 & 59.8 & 54.5 & 56.8 & 58.4 & 53.1 \\
Confidence        & 68.3 & 70.5 & 72.1 & 67.1 & 69.2 & 71.0 & 72.4 & 67.8 & 70.9 & 73.2 & 67.5 \\
Internal State Probing               & 66.2 & 68.4 & 70.5 & 64.8 & 66.7 & 68.1 & 69.5 & 65.5 & 69.1 & 71.3 & 65.4 \\
Self-Consistency     & 58.6 & 60.8 & 62.4 & 57.2 & 58.9 & 60.1 & 61.3 & 58.1 & 61.2 & 63.5 & 57.8 \\
Semantic Entropy     & 69.7 & 71.8 & 73.6 & 68.5 & 70.6 & 72.4 & 73.8 & 69.1 & 72.3 & 74.5 & 68.8 \\
\midrule
\textbf{Grad Detect (last 5)}  & 74.2 & 75.1 & 75.7 & 73.5 & 74.8 & 76.5 & 76.1 & 73.8 & 75.4 & 76.8 & 72.1 \\
\textbf{Grad Detect (all)}      & \textbf{75.0} & \textbf{76.2} & \textbf{76.8} & \textbf{74.3} & \textbf{75.5} & \textbf{76.9} & \textbf{77.4} & \textbf{74.6} & \textbf{76.5} & \textbf{77.8} & \textbf{73.2} \\
\bottomrule
\end{tabular}
\end{table*}

\section{Extended Baseline Analysis}
\label{app:baseline_analysis}

\paragraph{Self-assessment via prompting.}
The self-assessment baseline achieves only 49--54\% accuracy across
all models, performing worse than all other methods including simple
perplexity thresholding.
Models consistently overestimate their ability to answer correctly,
producing affirmative responses for approximately 78\% of queries
regardless of actual correctness.
This confirms that LLMs lack reliable introspective access to their
own knowledge boundaries~\cite{kadavath2022language} and that
external analysis of internal computation is necessary for reliable
detection.

\paragraph{Advantage over confidence baselines.}
The consistent advantage over confidence baselines
confirms that gradients expose decision-relevant internal structure
not visible in the output distribution.
A model can produce high-confidence hallucinations whose softmax
scores are indistinguishable from those of correct
answers~\cite{guo2017calibration}.
The gradients of such samples, however, point in measurably different
directions, as confirmed by the per-layer divergence analysis in
Section~\ref{sec:layer_ablation}.
This finding is consistent with observations that internal model
states encode truthfulness information beyond what output
probabilities
capture~\cite{azaria_mitchell_2023_llm_lying,su_2024_unsupervised_hallucination}.

\paragraph{Reference gradient sensitivity.}
Our default reference gradients are computed by averaging over all
samples in a category (Section~\ref{sec:ref_grad}).
We evaluated several alternatives.
Restricting to high-confidence samples degrades accuracy by
1--2 points. Restricting to uncertain samples degrades it by
6 points, as noisy gradients corrupt the prototype.
A stratified ensemble that maintains separate references per
confidence bin yields a marginal gain of 0.6 points at added
complexity.
We retain simple averaging as the default.
The same pattern holds for the Response task, where using all
abstention samples outperforms restricting to explicit refusals or
hedged responses.

\section{Extended Layer Ablation}
\label{app:layer_full}

\subsection{Layer Range Analysis}

Table~\ref{tab:layer_ablation} reports Correctness-task accuracy on
TriviaQA for seven layer configurations across all eleven models.
Trends are consistent across other datasets.

\begin{table*}[h]
\centering
\setlength{\tabcolsep}{3.8pt}
\footnotesize
\caption{Layer ablation (TriviaQA, Correctness task) across all
models.  We report classification accuracy (\%).  Performance
degrades gracefully when removing layers from either end, with a
maximum drop of 2--3 points for any contiguous subset covering at
least one-third of the network.}
\label{tab:layer_ablation}
\begin{tabular}{l ccc cccc ccc c}
\toprule
& \multicolumn{3}{c}{\textbf{Qwen2.5}}
& \multicolumn{4}{c}{\textbf{Falcon3}}
& \multicolumn{3}{c}{\textbf{Gemma-3}}
& \textbf{SmolLM3} \\
\cmidrule(lr){2-4}\cmidrule(lr){5-8}\cmidrule(lr){9-11}\cmidrule(lr){12-12}
\textbf{Layer Range} & 1.5B & 3B & 7B & 1B & 3B & 7B & 10B & 1B & 4B & 12B & 3B \\
\midrule
All layers       & 75.0 & 76.2 & 76.8 & 74.3 & 75.5 & 76.9 & 77.4 & 74.6 & 76.5 & 77.8 & 73.2 \\
\addlinespace
Last 20          & 74.8 & 76.0 & 76.6 & 74.1 & 75.3 & 76.7 & 77.2 & 74.4 & 76.3 & 77.6 & 73.0 \\
Last 10          & 74.5 & 75.7 & 76.3 & 73.8 & 75.0 & 76.4 & 76.9 & 74.1 & 76.0 & 77.3 & 72.7 \\
Last 5           & 74.2 & 75.1 & 75.7 & 73.5 & 74.8 & 76.5 & 76.1 & 73.8 & 75.4 & 76.8 & 72.1 \\
\addlinespace
First 5          & 72.8 & 74.0 & 74.6 & 72.1 & 73.3 & 74.8 & 75.3 & 72.5 & 74.3 & 75.6 & 71.0 \\
First 10         & 73.4 & 74.6 & 75.2 & 72.7 & 73.9 & 75.4 & 75.9 & 73.1 & 74.9 & 76.2 & 71.6 \\
First 20         & 74.3 & 75.5 & 76.1 & 73.6 & 74.8 & 76.2 & 76.7 & 73.9 & 75.8 & 77.1 & 72.5 \\
\bottomrule
\end{tabular}
\end{table*}

\paragraph{Discriminative information is distributed but concentrates
in later layers.}
Across all eleven models, the last 5 layers retain 98--99\% of
full-model accuracy, dropping only 0.5--1.0 points on average.
The last 10 and last 20 layers lose even less (0.4--0.7 and
0.2--0.3 points respectively).
Conversely, the first 5 layers alone still achieve 96--97\% of
full accuracy, indicating that even early layers carry a meaningful
gradient signal.
Adding more early layers improves performance monotonically. The
first 10 layers recover 97--98\% and the first 20 layers recover
98--99\% of full accuracy.
The maximum drop for any configuration covering at least one-third
of the network is 2--3 points, confirming that the discriminative
gradient signal is broadly distributed across the transformer.
Nevertheless, later layers consistently outperform earlier layers
of the same width. For example, last 5 vs.\ first 5 differs by 1.2--1.4
points. This aligns with the established view that later transformer
layers handle high-level semantic
integration~\cite{tenney2019bert,jawahar2019does} where factual
grounding succeeds or fails.

\paragraph{Layers are complementary.}
Although any contiguous subset of layers achieves strong accuracy,
combining layers from different regions of the network yields
consistent gains.
Self-attention in the predictor exploits cross-layer patterns that
no single region captures in isolation.
This is consistent with circuit-level analyses showing that
transformer computations emerge from interactions across multiple
layers~\cite{elhage2021mathematical}.

\section{Inference Time Breakdown}
\label{app:timing}

Table~\ref{tab:inference_time} reports end-to-end inference time per
sample for all detection methods, measured on Qwen2.5-3B with a
single NVIDIA A100 (40\,GB).

Grad Detect with all layers adds 47\,ms to the standard forward pass
(45\,ms), for a total of 92\,ms ($2.04\times$). This overhead has
three components: the backward pass (38\,ms), cosine
similarity computation against reference gradients (3\,ms), and the
predictor's forward pass (6\,ms). The backward pass dominates because
it requires propagating gradients through the full model.

Using only the last 5 layers reduces total time to 68\,ms by computing gradients for a subset of parameters.
This is substantially cheaper than Self-Consistency and
Semantic Entropy, which both require multiple full
forward passes through the LLM.

Confidence baselines add negligible overhead (2\,ms)
because they require only reading out statistics from the existing
forward pass output distribution.

\begin{table}[h]
\centering\small
\begin{tabular}{lcc}
\toprule
Method & Time (ms) \\
\midrule
Standard forward pass          &  45  \\
\midrule
+ Confidence                   &  47  \\
+ Grad Detect (last 5 layers)  &  68  \\
+ Grad Detect (all layers)     &  92  \\
\midrule
+ Self-Consistency (5 gens)    & 225 \\
+ Semantic Entropy (10 gens)   & 450  \\
\bottomrule
\end{tabular}
\caption{End-to-end inference time per sample (Qwen2.5-3B, A100).
Grad Detect overhead comprises: backward pass (38\,ms), cosine
similarity (3\,ms), and predictor forward pass (6\,ms).}
\label{tab:inference_time}
\end{table}

\section{Gradient Signature Visualization}
\label{app:lda_plots}

We apply Linear Discriminant Analysis (LDA) to the layer-wise
cosine similarity vectors to visualize the separability of
gradient signatures across behavioral categories. For each model,
every point in the resulting projection represents a single
sample, positioned according to its cosine similarity profile
across layers and reference gradients. Clear separation between
category clusters provides visual confirmation that gradient
directions carry discriminative information exploitable by the
transformer-based predictor.

Visualizations are organized along three dimensions.
\textbf{Task} determines the classification objective: Full
(three-way), Correctness (binary), or Response (binary).
\textbf{Reference class} indicates which category prototype
(Correct, Incorrect, or Did Not Answer) serves as the comparison
anchor in the cosine similarity computation.
\textbf{Response type} specifies which probe response
(affirming $r^{+}$ or rejection $r^{-}$, as defined in
Section~\ref{sec:grad_extraction}) is used to compute gradients
for the reference samples and training samples respectively.
All plots show results for Qwen2.5-3B, Falcon3-3B, Gemma3-4B, and
Gemma3-1B.

\subsection{Full Task (Correct vs.\ Incorrect vs.\ Did Not Answer)}
\label{app:lda_full}

\begin{figure*}[htbp]
  \centering
  \begin{subfigure}[b]{0.48\textwidth}
    \centering
    \includegraphics[width=\linewidth]{figures/lda_plots/Full/compare_correct_affirming_affirming.png}
    \caption{Ref: affirming, Train: affirming}
  \end{subfigure}
  \hfill
  \begin{subfigure}[b]{0.48\textwidth}
    \centering
    \includegraphics[width=\linewidth]{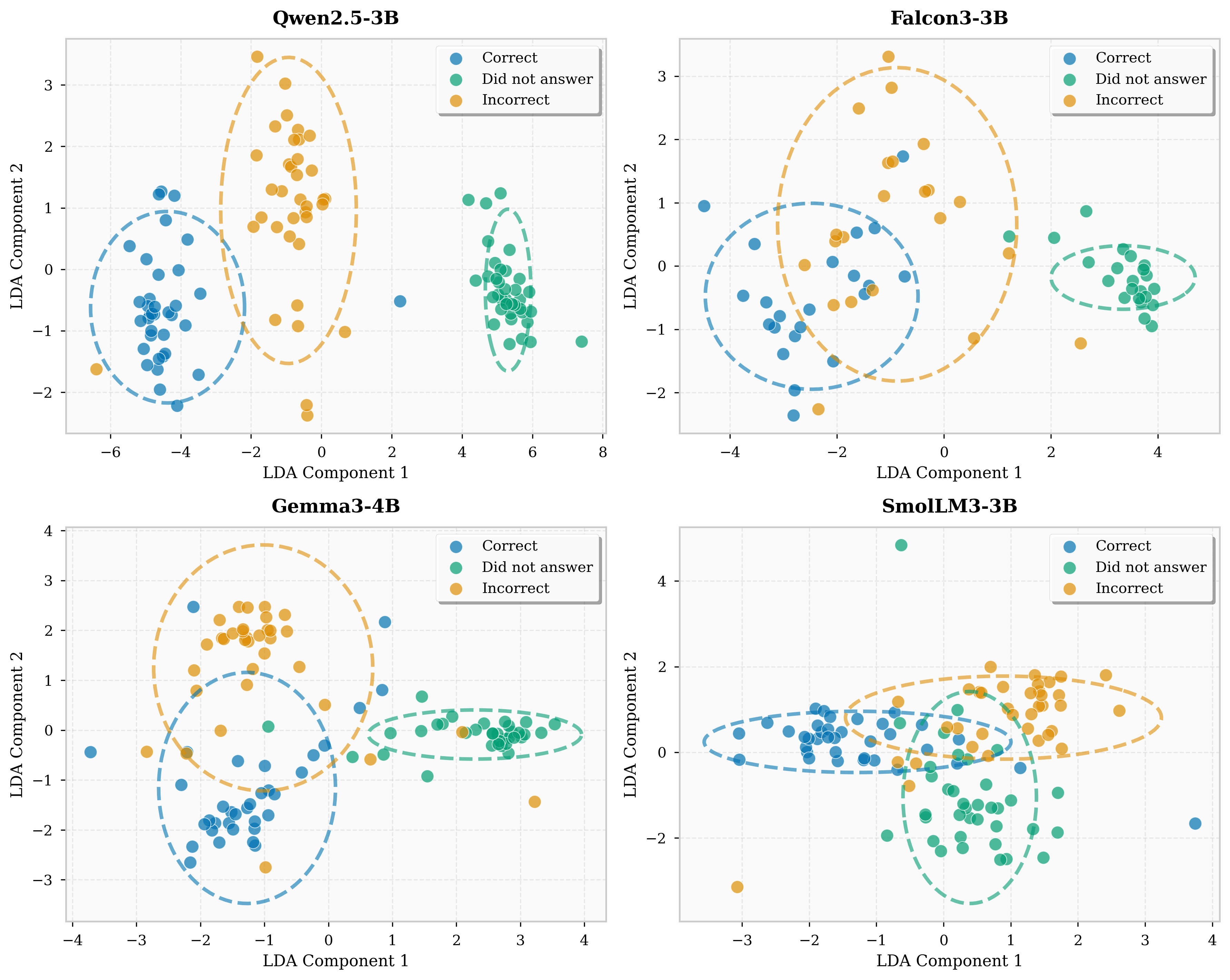}
    \caption{Ref: affirming, Train: rejection}
  \end{subfigure}

  \medskip

  \begin{subfigure}[b]{0.48\textwidth}
    \centering
    \includegraphics[width=\linewidth]{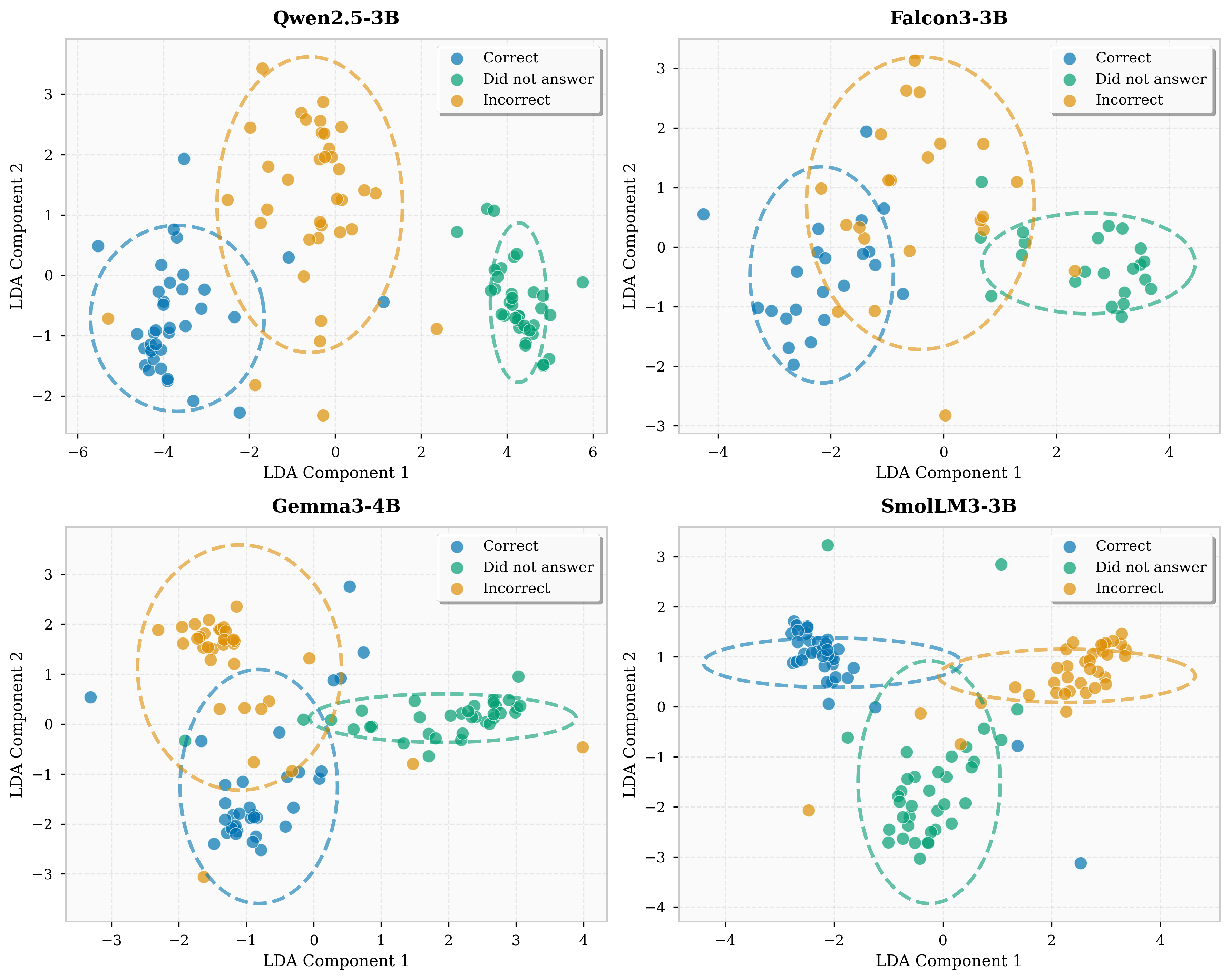}
    \caption{Ref: rejection, Train: affirming}
  \end{subfigure}
  \hfill
  \begin{subfigure}[b]{0.48\textwidth}
    \centering
    \includegraphics[width=\linewidth]{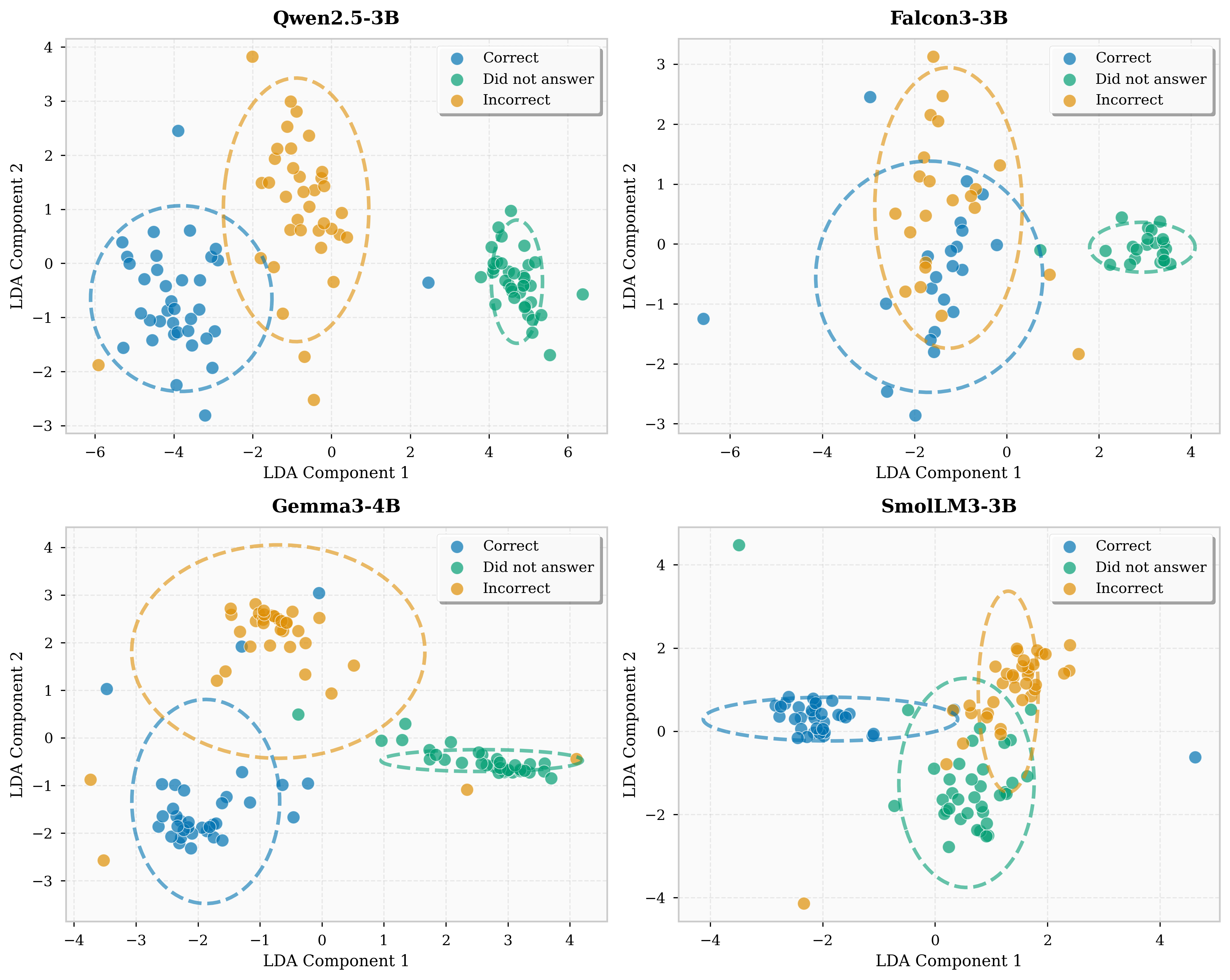}
    \caption{Ref: rejection, Train: rejection}
  \end{subfigure}
  \caption{Full task --- \textbf{Correct} reference gradient. LDA
  projections of layer-wise cosine similarities for all four
  combinations of reference and training response types.}
  \label{fig:lda_full_correct}
\end{figure*}

\begin{figure*}[htbp]
  \centering
  \begin{subfigure}[b]{0.48\textwidth}
    \centering
    \includegraphics[width=\linewidth]{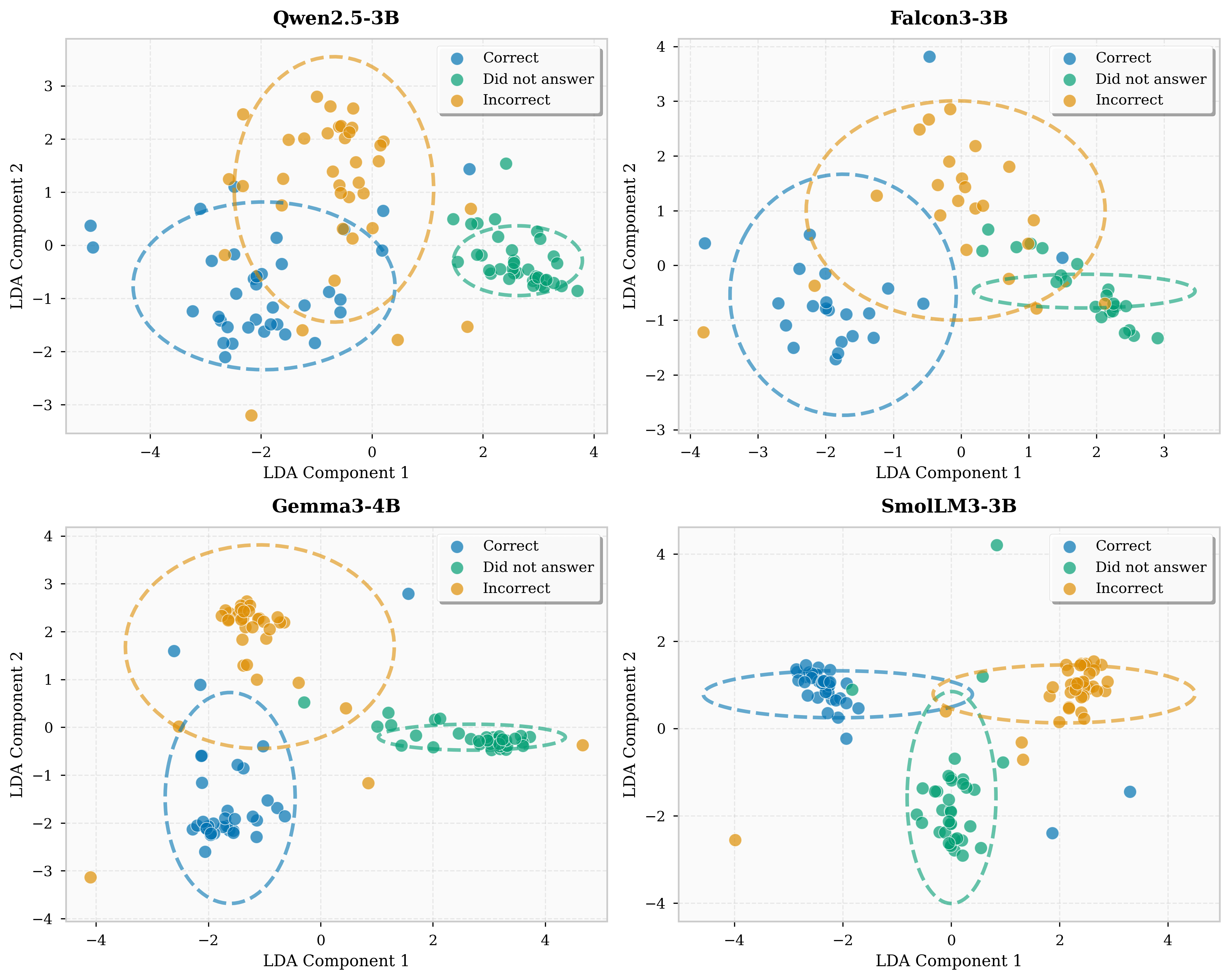}
    \caption{Ref: affirming, Train: affirming}
  \end{subfigure}
  \hfill
  \begin{subfigure}[b]{0.48\textwidth}
    \centering
    \includegraphics[width=\linewidth]{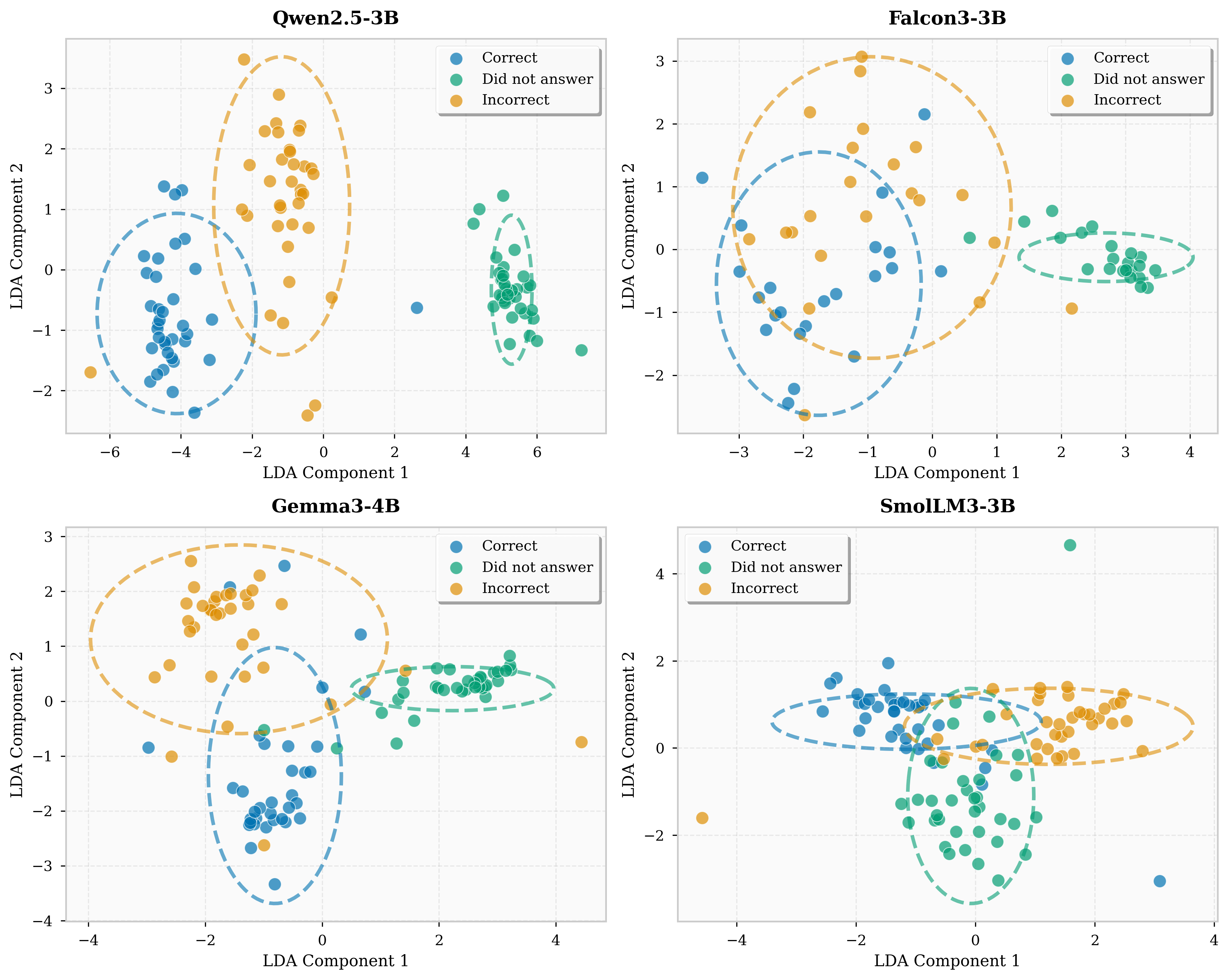}
    \caption{Ref: affirming, Train: rejection}
  \end{subfigure}

  \medskip

  \begin{subfigure}[b]{0.48\textwidth}
    \centering
    \includegraphics[width=\linewidth]{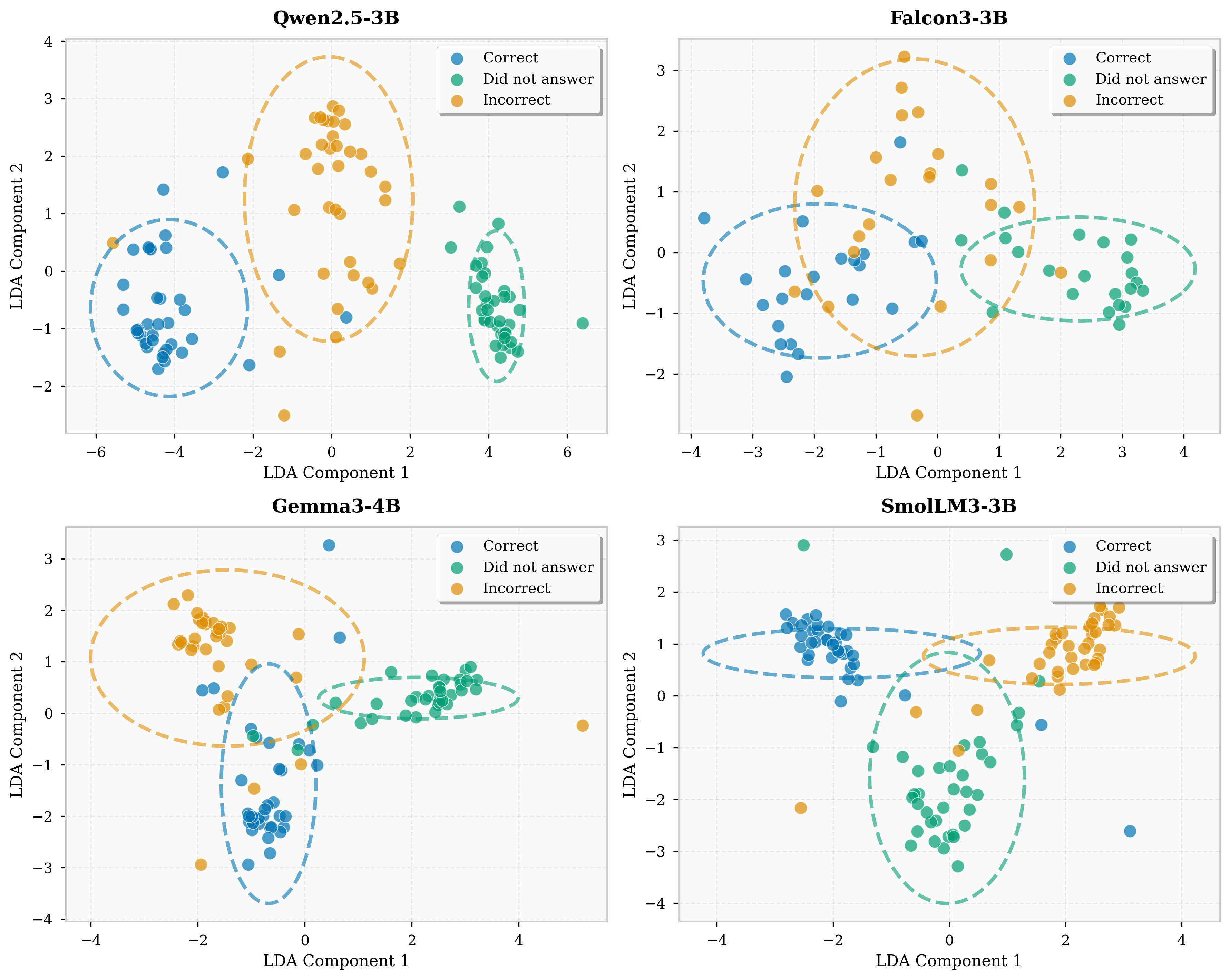}
    \caption{Ref: rejection, Train: affirming}
  \end{subfigure}
  \hfill
  \begin{subfigure}[b]{0.48\textwidth}
    \centering
    \includegraphics[width=\linewidth]{figures/lda_plots/Full/compare_correct_rejection_rejection.png}
    \caption{Ref: rejection, Train: rejection}
  \end{subfigure}
  \caption{Full task --- \textbf{Incorrect} reference gradient.
  LDA projections of layer-wise cosine similarities for all four
  combinations of reference and training response types.}
  \label{fig:lda_full_incorrect}
\end{figure*}

\begin{figure*}[htbp]
  \centering
  \begin{subfigure}[b]{0.48\textwidth}
    \centering
    \includegraphics[width=\linewidth]{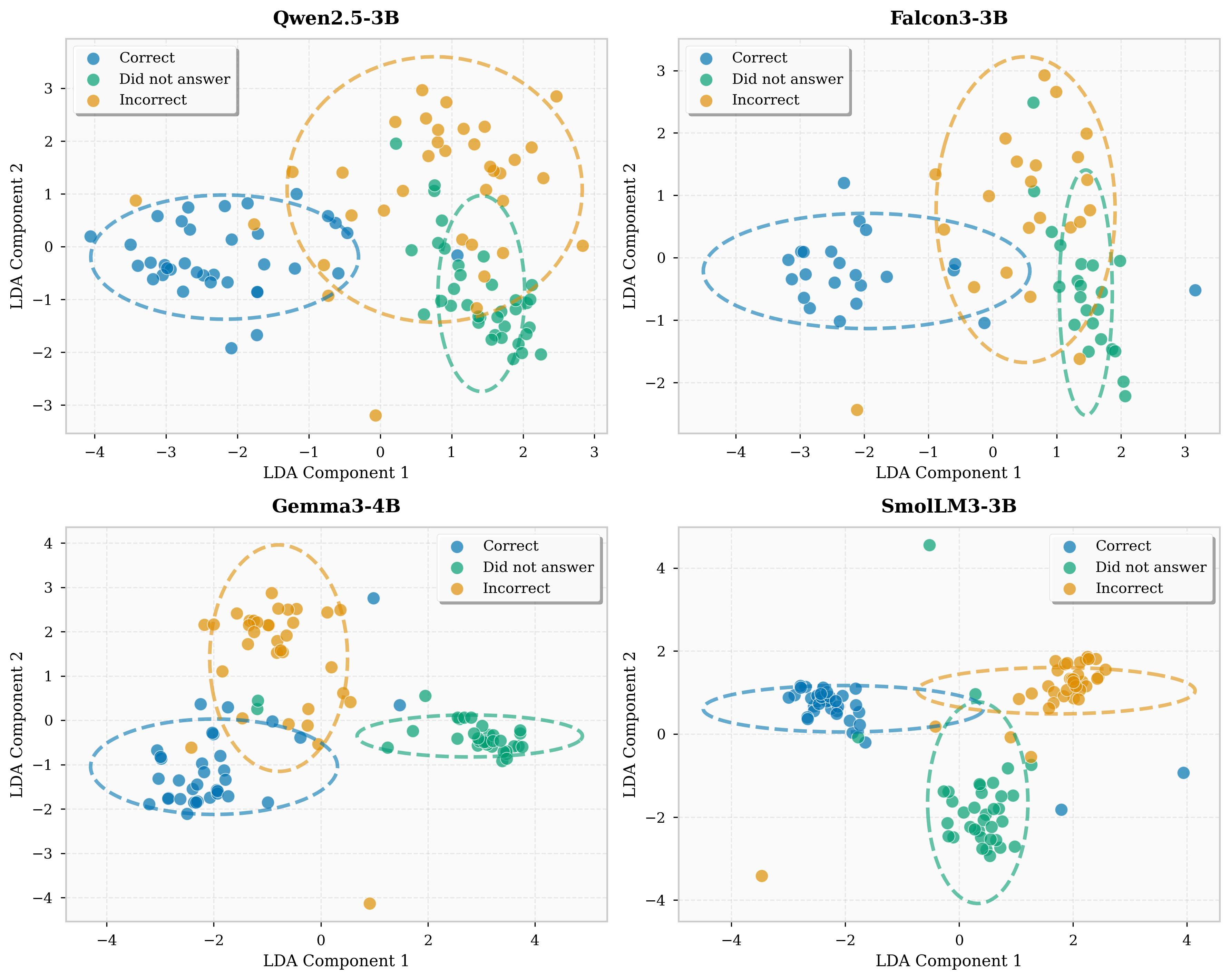}
    \caption{Ref: affirming, Train: affirming}
  \end{subfigure}
  \hfill
  \begin{subfigure}[b]{0.48\textwidth}
    \centering
    \includegraphics[width=\linewidth]{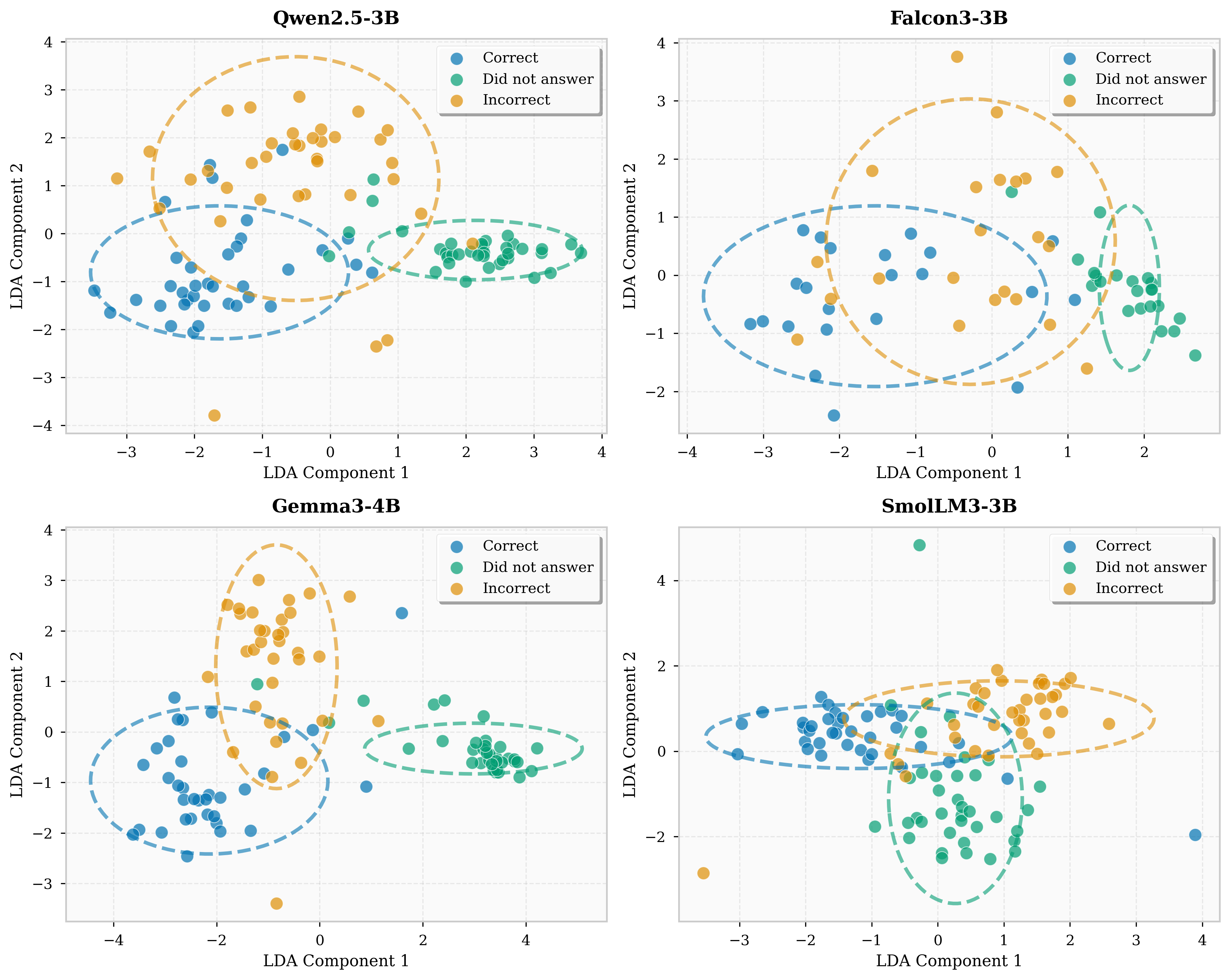}
    \caption{Ref: affirming, Train: rejection}
  \end{subfigure}

  \medskip

  \begin{subfigure}[b]{0.48\textwidth}
    \centering
    \includegraphics[width=\linewidth]{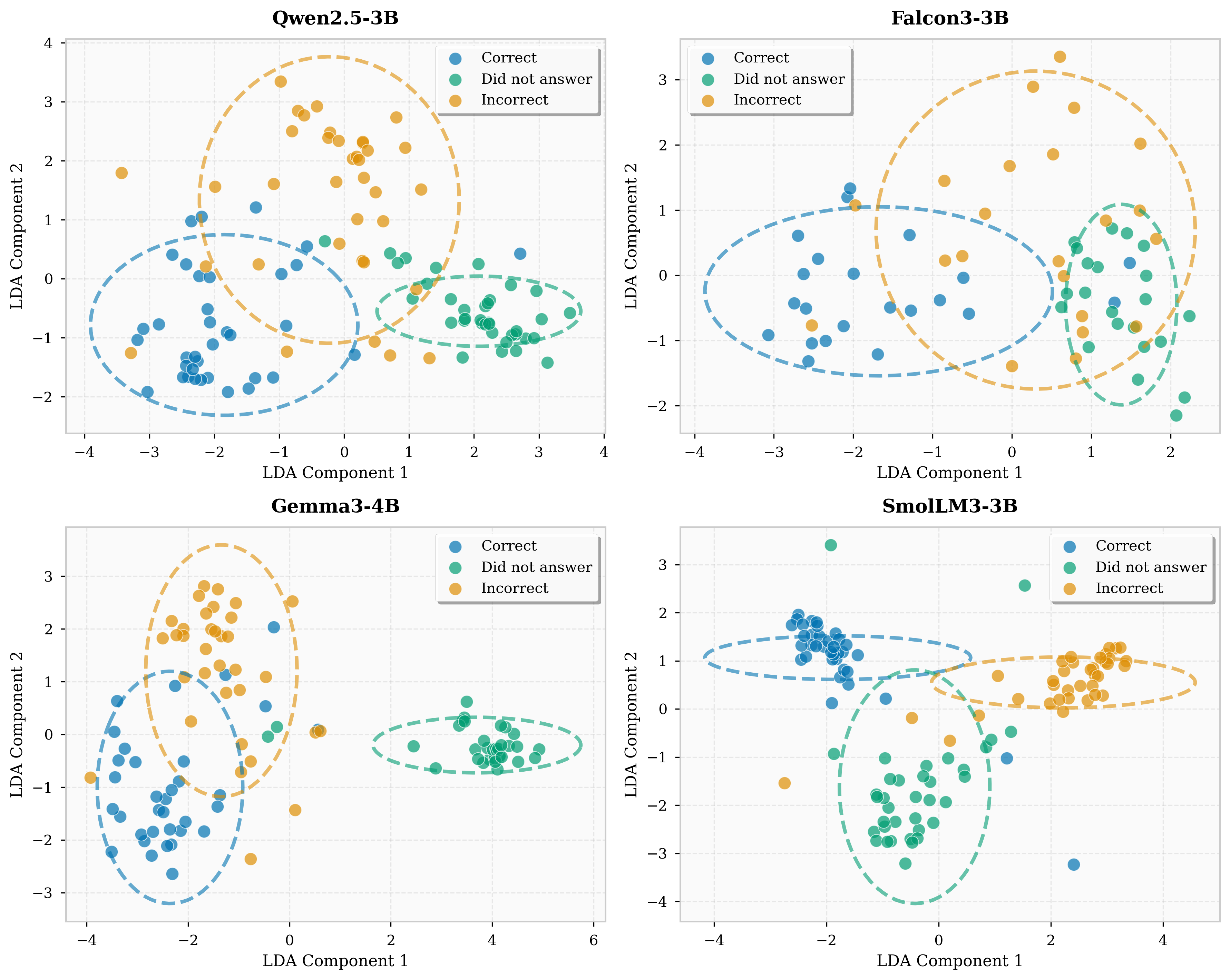}
    \caption{Ref: rejection, Train: affirming}
  \end{subfigure}
  \hfill
  \begin{subfigure}[b]{0.48\textwidth}
    \centering
    \includegraphics[width=\linewidth]{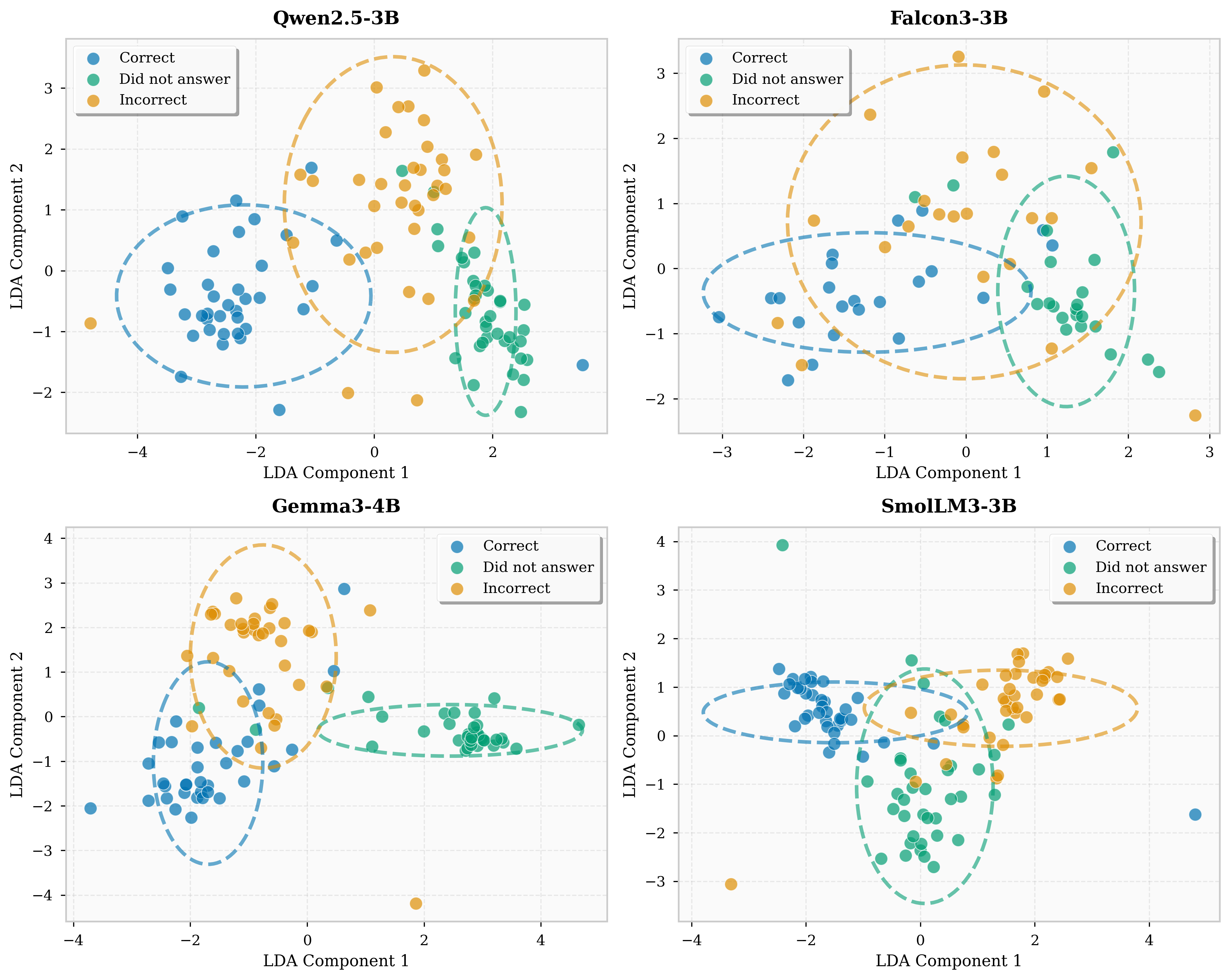}
    \caption{Ref: rejection, Train: rejection}
  \end{subfigure}
  \caption{Full task --- \textbf{Did Not Answer} reference
  gradient. LDA projections of layer-wise cosine similarities for
  all four combinations of reference and training response types.}
  \label{fig:lda_full_dna}
\end{figure*}

Figures~\ref{fig:lda_full_correct}--\ref{fig:lda_full_dna} show
LDA projections for the three-way Full task, grouped by reference
class. Across all response-type combinations, the three behavioral
categories form distinct clusters. Separation is strongest in the
affirming/affirming setting (subfigure (a) in each figure), where
both reference samples and training samples come from substantive
responses. The rejection/rejection setting (subfigure (d)) also
shows clear separation, confirming that gradient signatures remain
discriminative regardless of whether the model answered or
abstained.

\subsection{Correctness Task (Correct vs.\ Incorrect)}
\label{app:lda_correctness}

\begin{figure*}[htbp]
  \centering
  \begin{subfigure}[b]{0.48\textwidth}
    \centering
    \includegraphics[width=\linewidth]{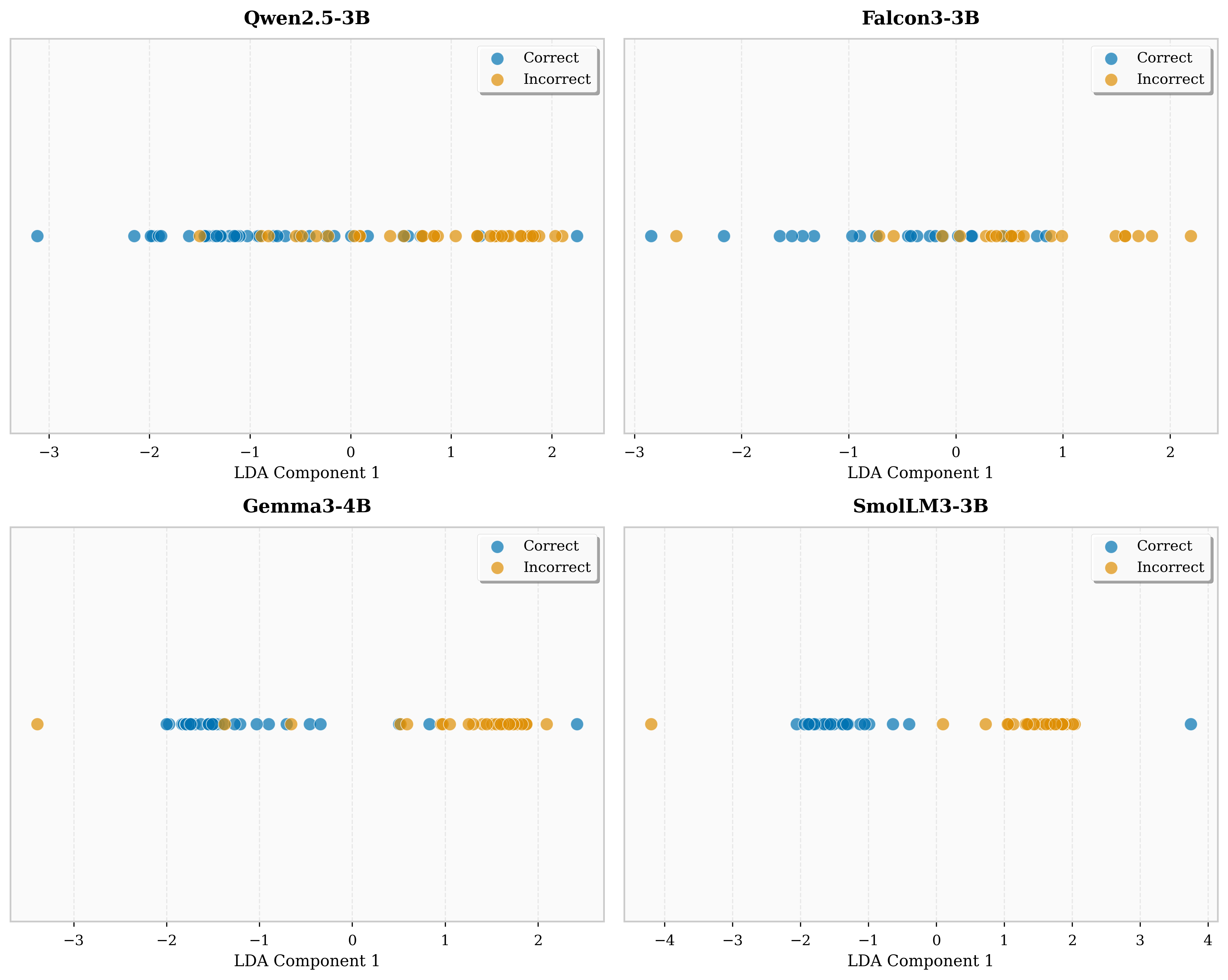}
    \caption{Ref: affirming, Train: affirming}
  \end{subfigure}
  \hfill
  \begin{subfigure}[b]{0.48\textwidth}
    \centering
    \includegraphics[width=\linewidth]{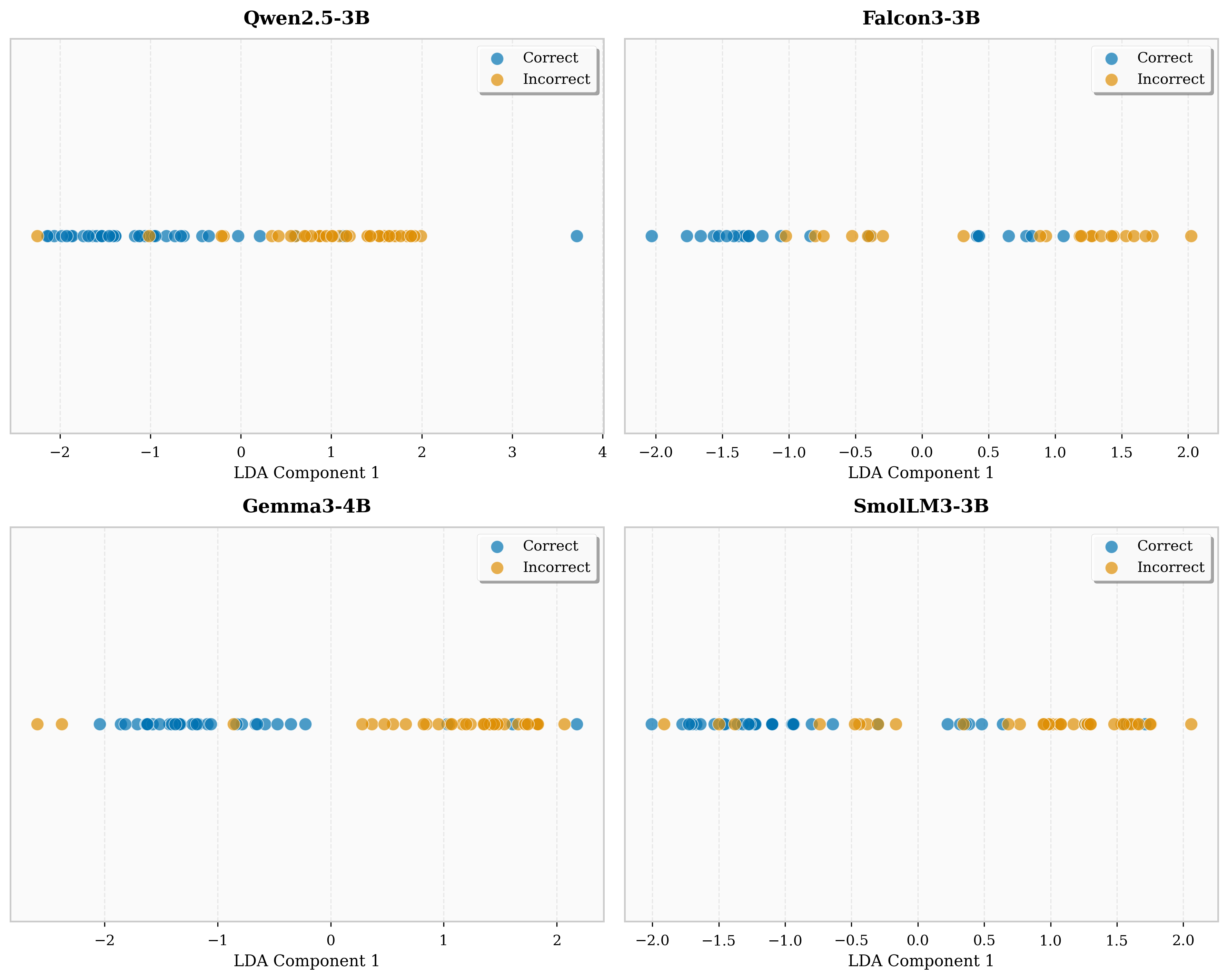}
    \caption{Ref: affirming, Train: rejection}
  \end{subfigure}

  \medskip

  \begin{subfigure}[b]{0.48\textwidth}
    \centering
    \includegraphics[width=\linewidth]{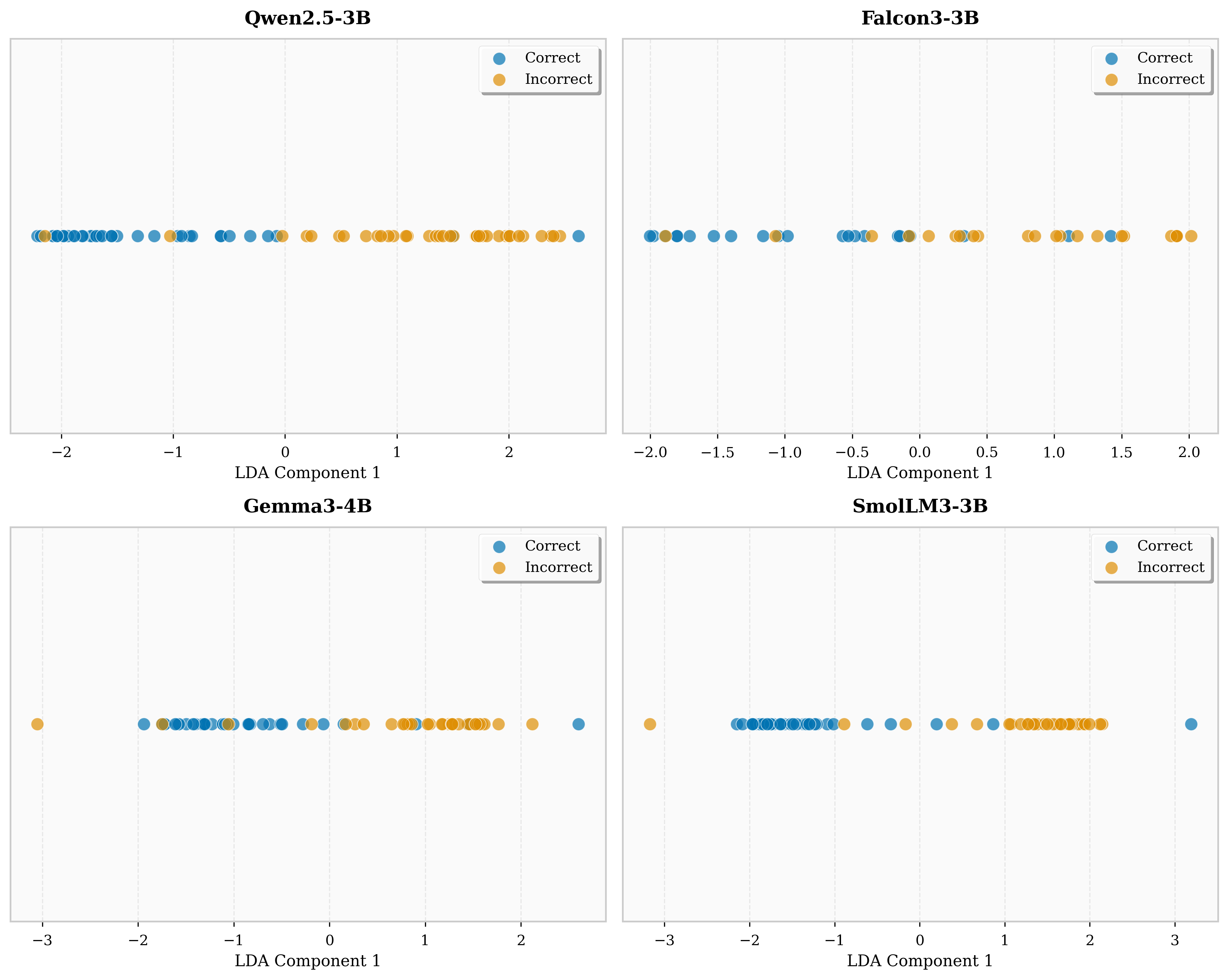}
    \caption{Ref: rejection, Train: affirming}
  \end{subfigure}
  \hfill
  \begin{subfigure}[b]{0.48\textwidth}
    \centering
    \includegraphics[width=\linewidth]{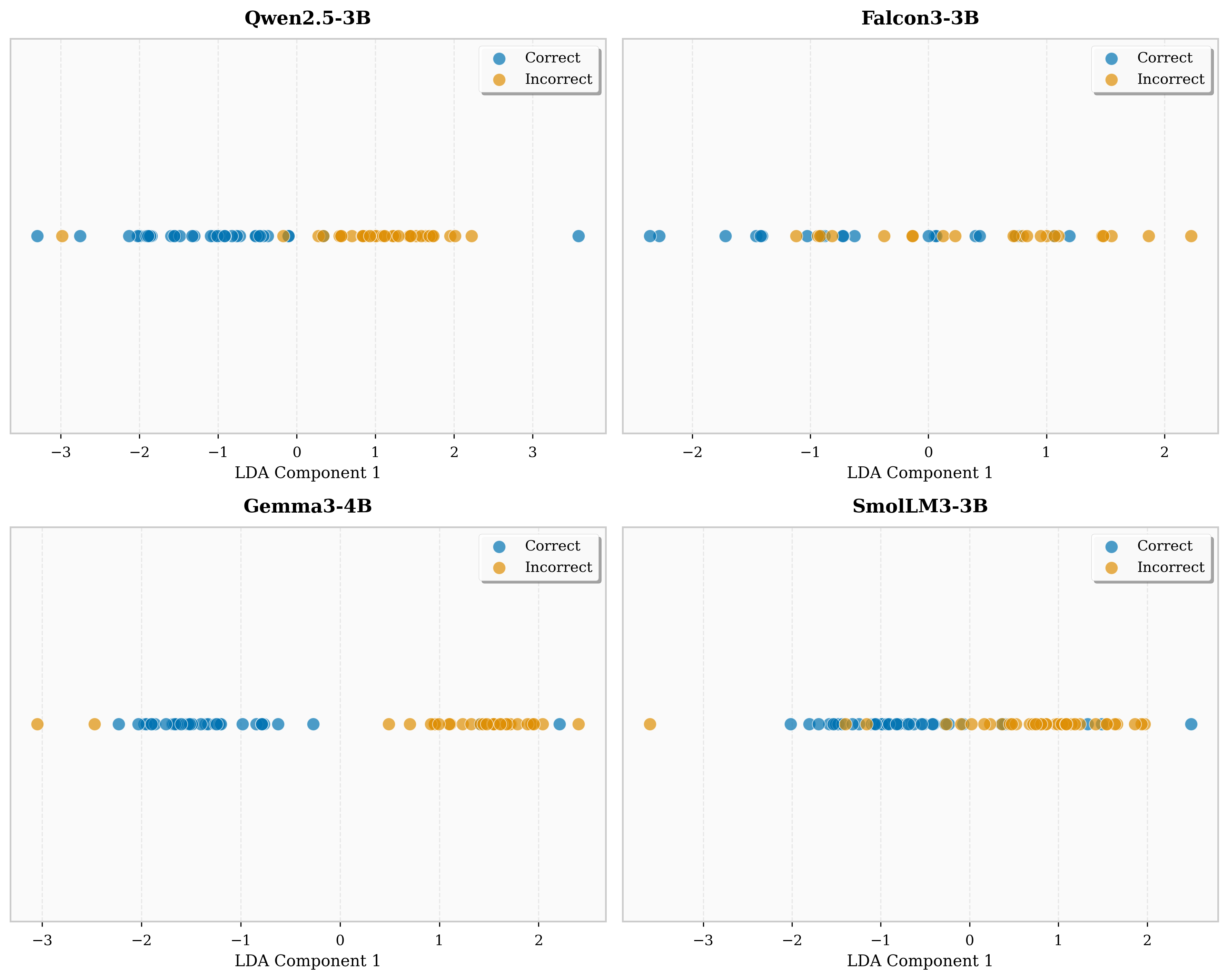}
    \caption{Ref: rejection, Train: rejection}
  \end{subfigure}
  \caption{Correctness task --- \textbf{Correct} reference
  gradient. LDA projections of layer-wise cosine similarities.}
  \label{fig:lda_corr_correct}
\end{figure*}

\begin{figure*}[htbp]
  \centering
  \begin{subfigure}[b]{0.48\textwidth}
    \centering
    \includegraphics[width=\linewidth]{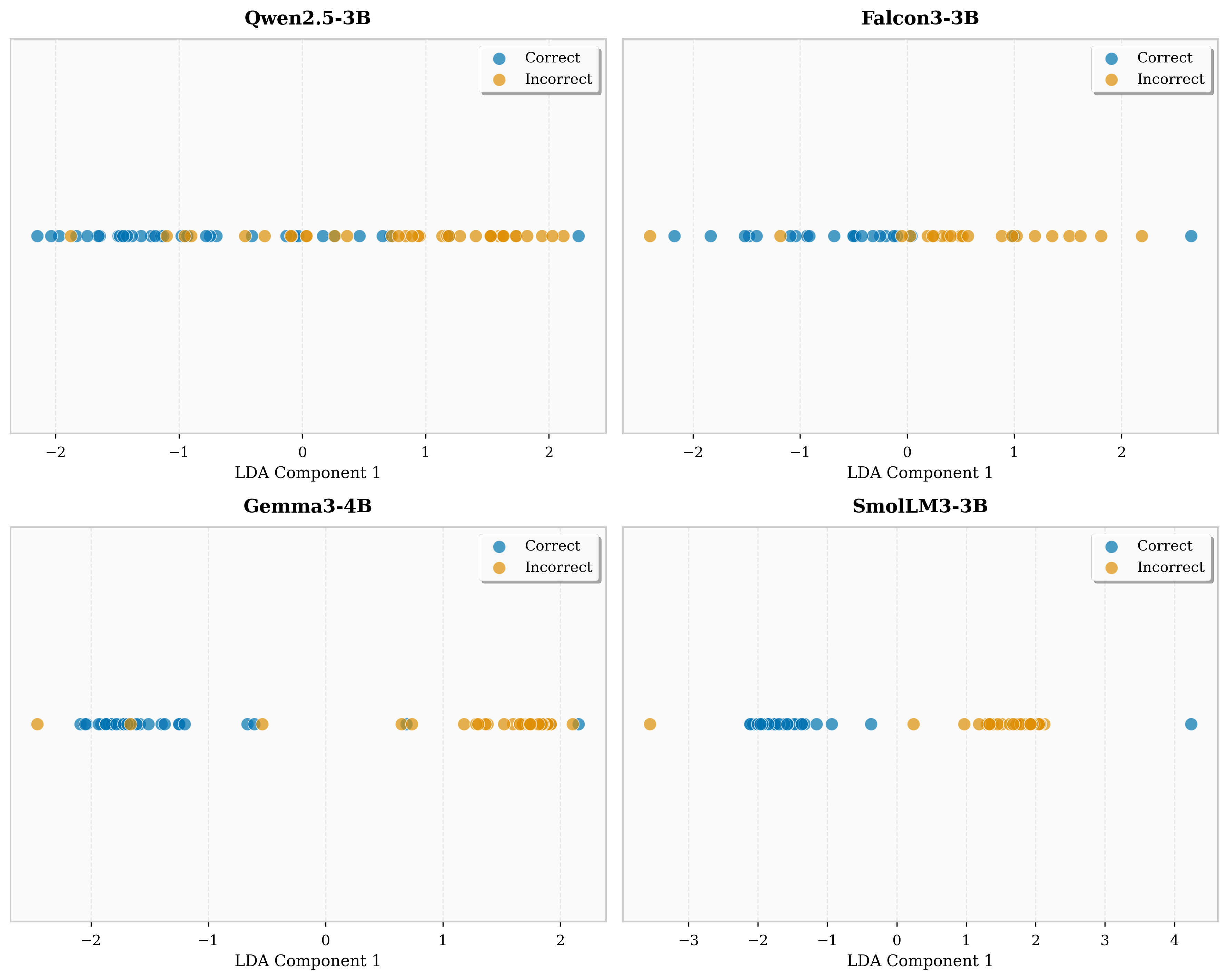}
    \caption{Ref: affirming, Train: affirming}
  \end{subfigure}
  \hfill
  \begin{subfigure}[b]{0.48\textwidth}
    \centering
    \includegraphics[width=\linewidth]{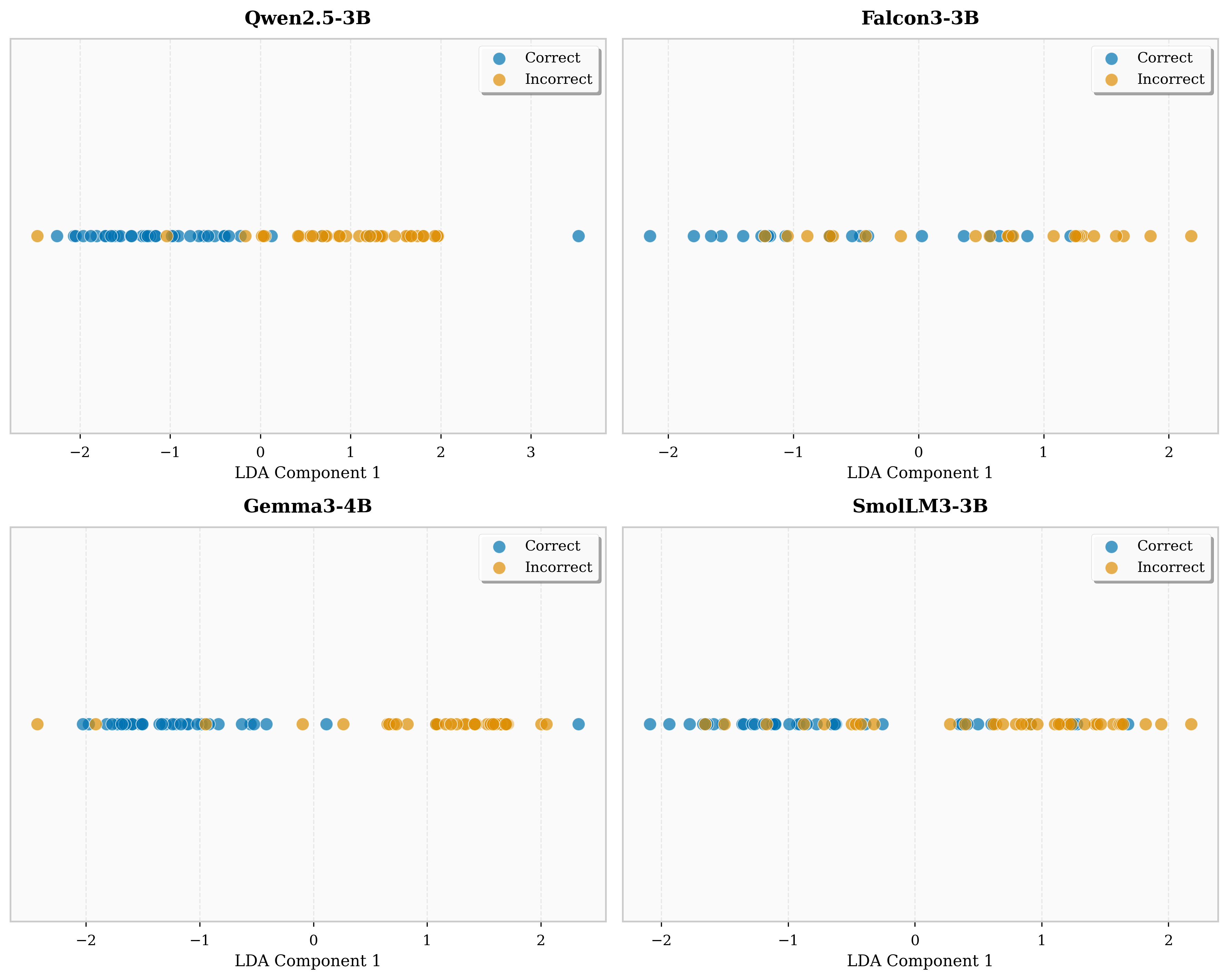}
    \caption{Ref: affirming, Train: rejection}
  \end{subfigure}

  \medskip

  \begin{subfigure}[b]{0.48\textwidth}
    \centering
    \includegraphics[width=\linewidth]{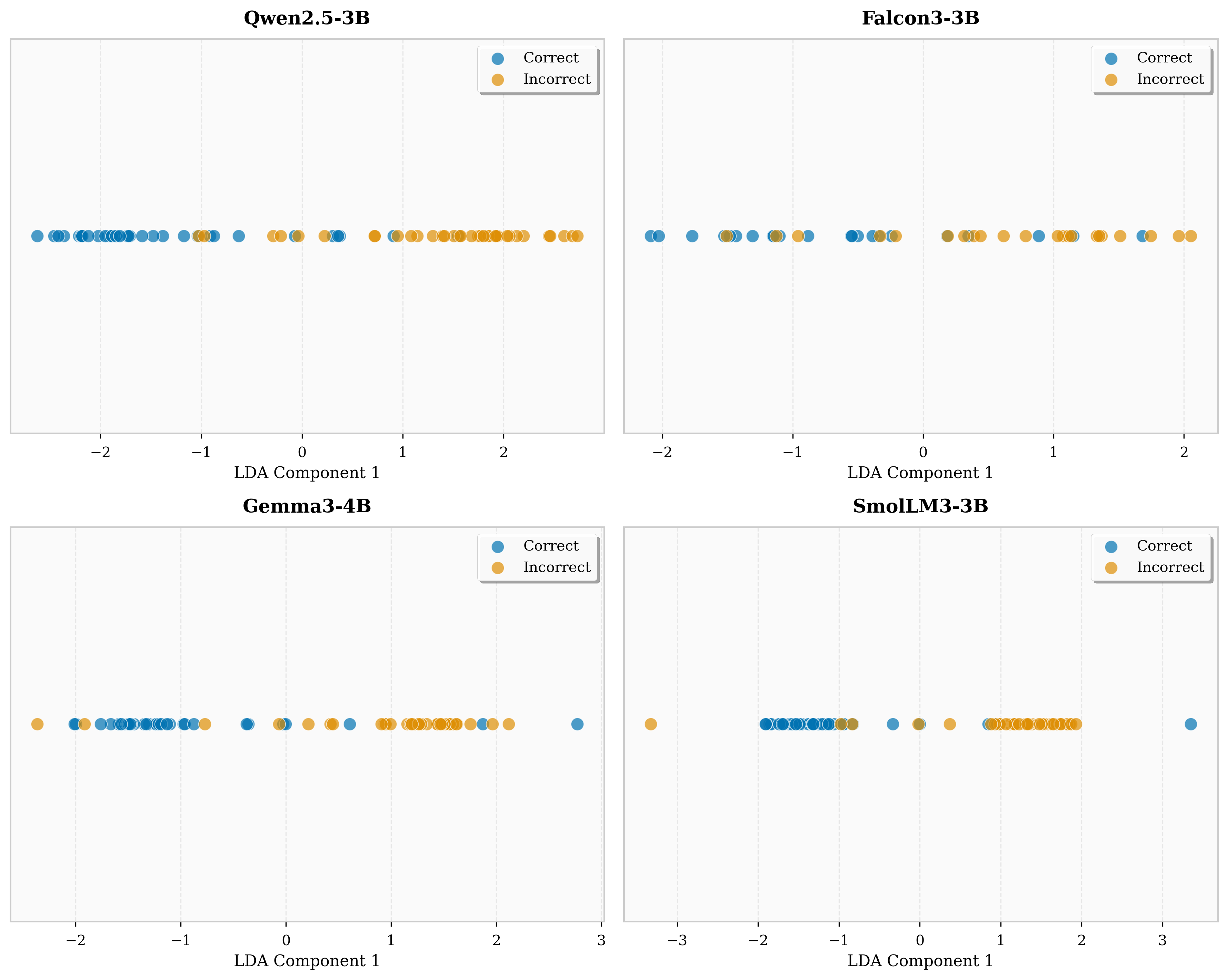}
    \caption{Ref: rejection, Train: affirming}
  \end{subfigure}
  \hfill
  \begin{subfigure}[b]{0.48\textwidth}
    \centering
    \includegraphics[width=\linewidth]{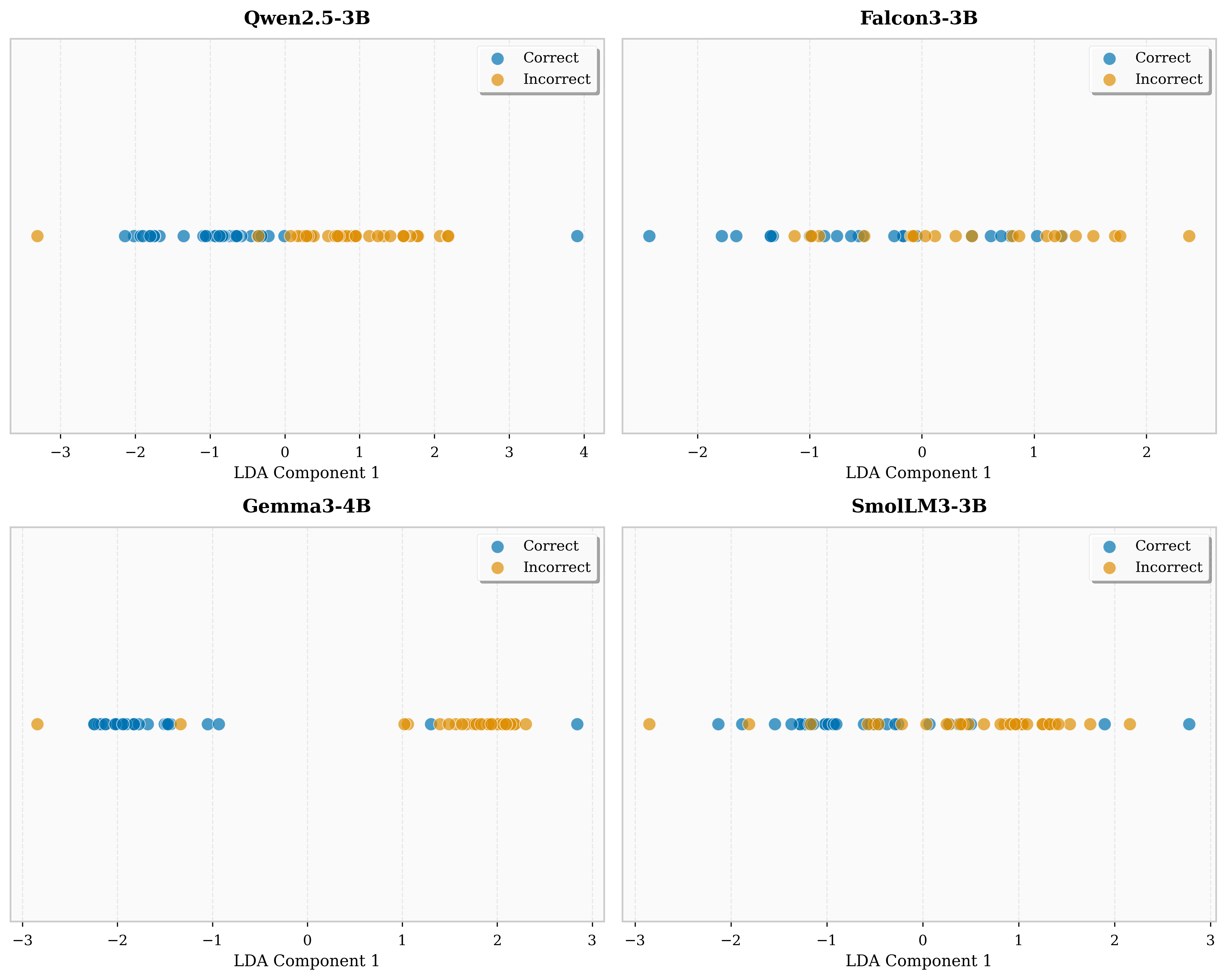}
    \caption{Ref: rejection, Train: rejection}
  \end{subfigure}
  \caption{Correctness task --- \textbf{Incorrect} reference
  gradient. LDA projections of layer-wise cosine similarities.}
  \label{fig:lda_corr_incorrect}
\end{figure*}

Figures~\ref{fig:lda_corr_correct}
and~\ref{fig:lda_corr_incorrect} show LDA projections for the
binary Correctness task. Although Did Not Answer samples are
excluded from classification, the cosine similarity features are
still computed against all three reference gradients. The
separation between Correct and Incorrect clusters is tighter than
in the Full task, reflecting the greater difficulty of this binary
distinction.

\subsection{Response Task (Answered vs.\ Did Not Answer)}
\label{app:lda_response}

\begin{figure*}[htbp]
  \centering
  \begin{subfigure}[b]{0.48\textwidth}
    \centering
    \includegraphics[width=\linewidth]{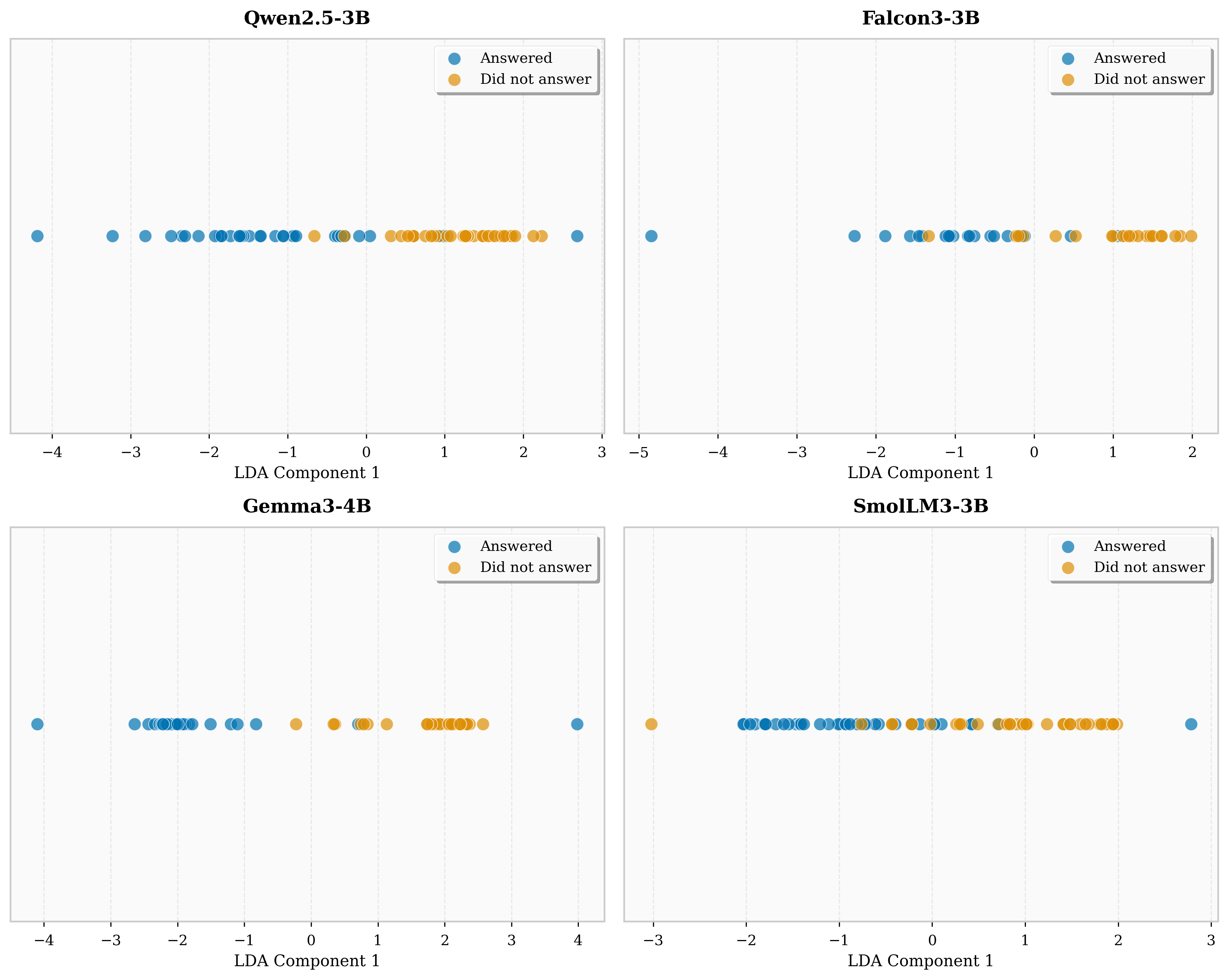}
    \caption{Ref: affirming, Train: affirming}
  \end{subfigure}
  \hfill
  \begin{subfigure}[b]{0.48\textwidth}
    \centering
    \includegraphics[width=\linewidth]{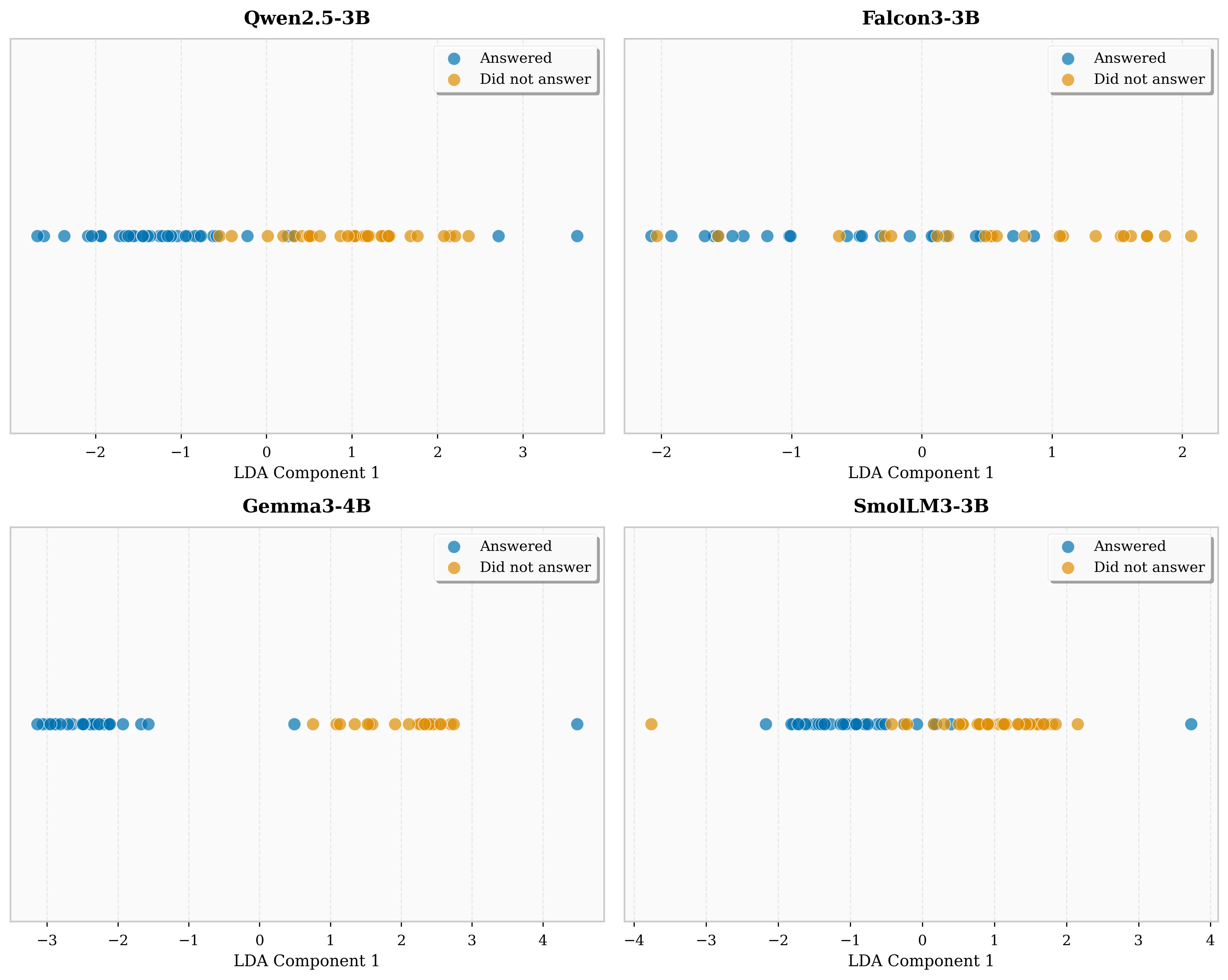}
    \caption{Ref: affirming, Train: rejection}
  \end{subfigure}

  \medskip

  \begin{subfigure}[b]{0.48\textwidth}
    \centering
    \includegraphics[width=\linewidth]{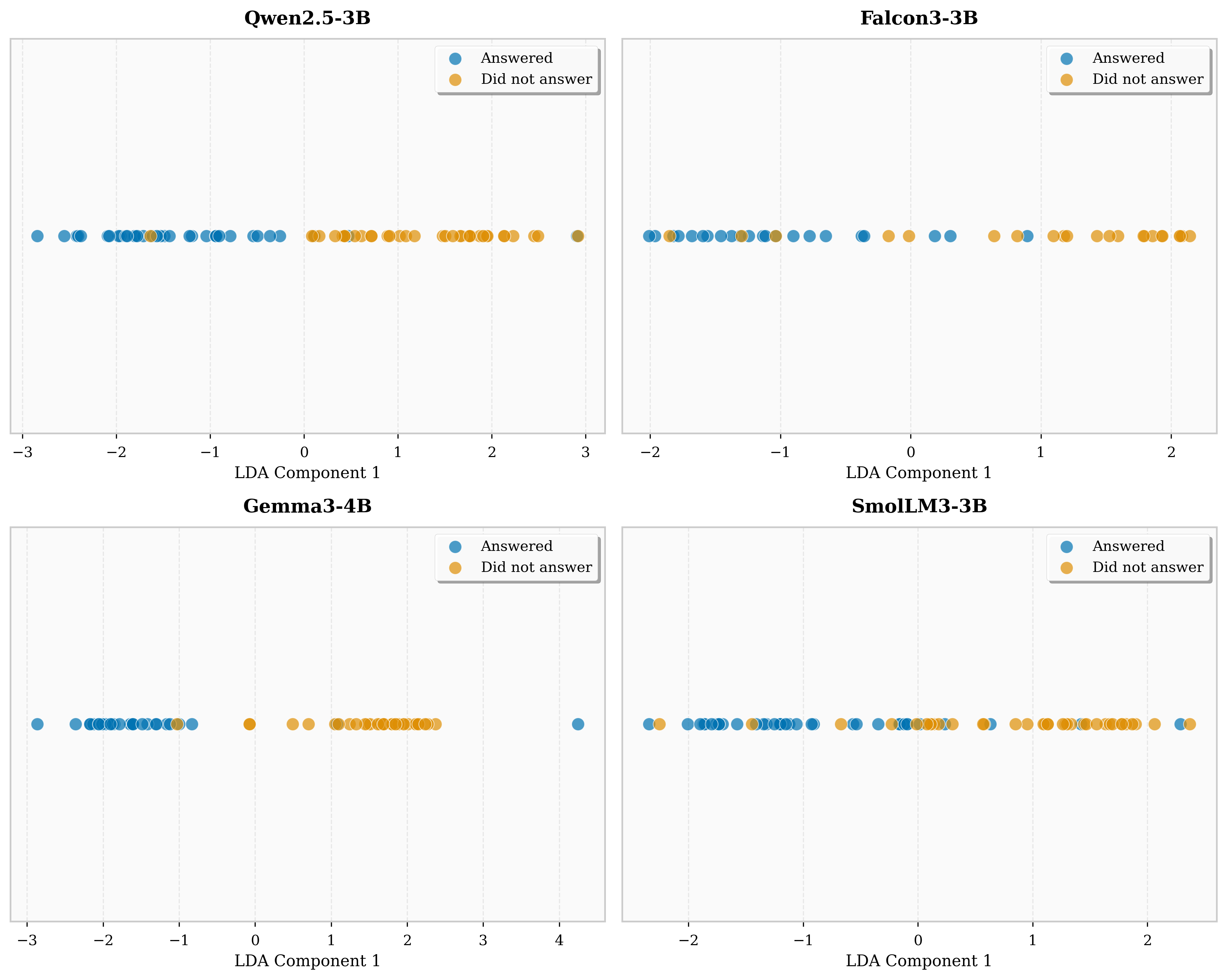}
    \caption{Ref: rejection, Train: affirming}
  \end{subfigure}
  \hfill
  \begin{subfigure}[b]{0.48\textwidth}
    \centering
    \includegraphics[width=\linewidth]{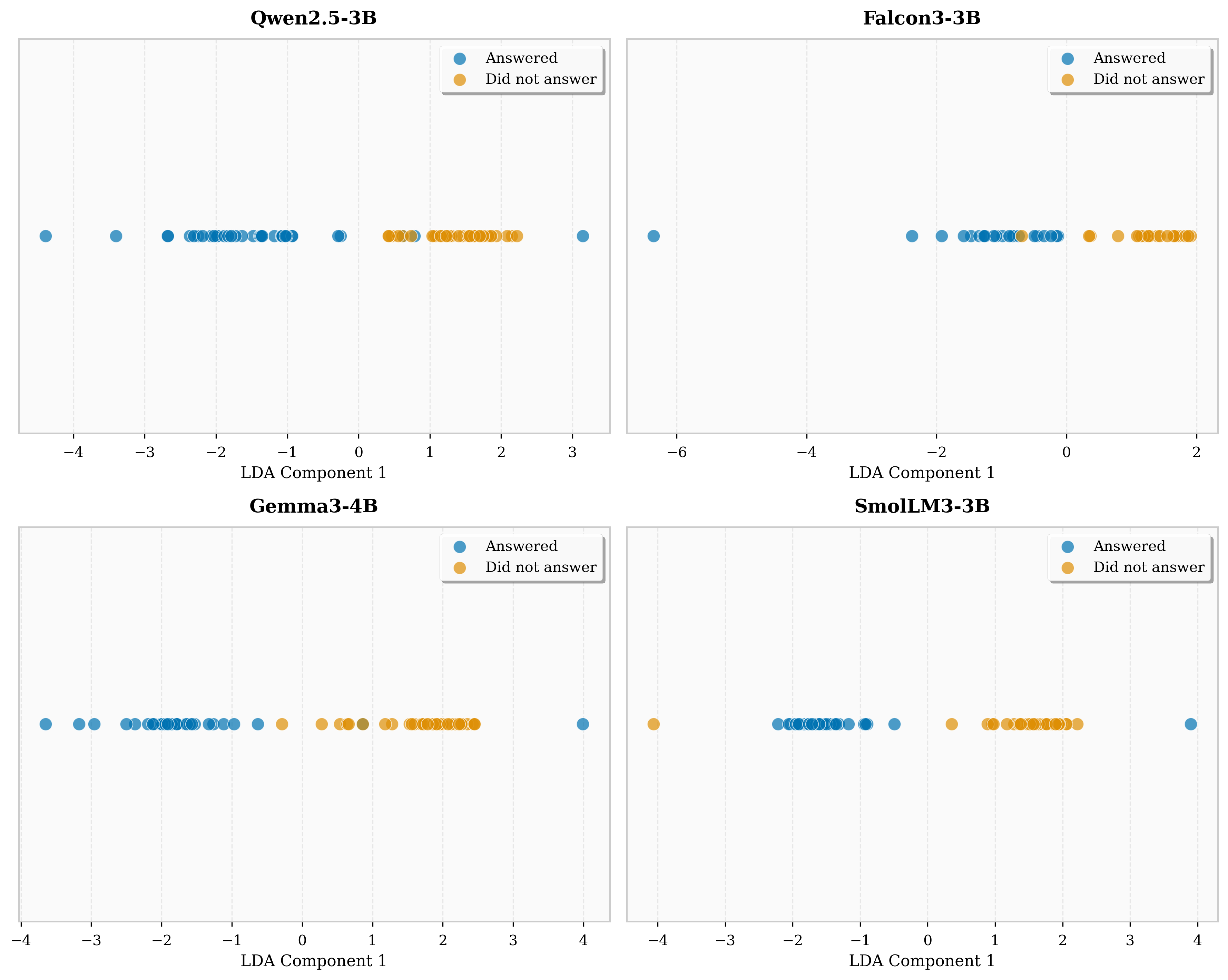}
    \caption{Ref: rejection, Train: rejection}
  \end{subfigure}
  \caption{Response task --- \textbf{Answered} reference gradient.
  LDA projections of layer-wise cosine similarities. Wide cluster
  separation reflects the near-perfect accuracy of abstention
  detection.}
  \label{fig:lda_resp_correct}
\end{figure*}

\begin{figure*}[htbp]
  \centering
  \begin{subfigure}[b]{0.48\textwidth}
    \centering
    \includegraphics[width=\linewidth]{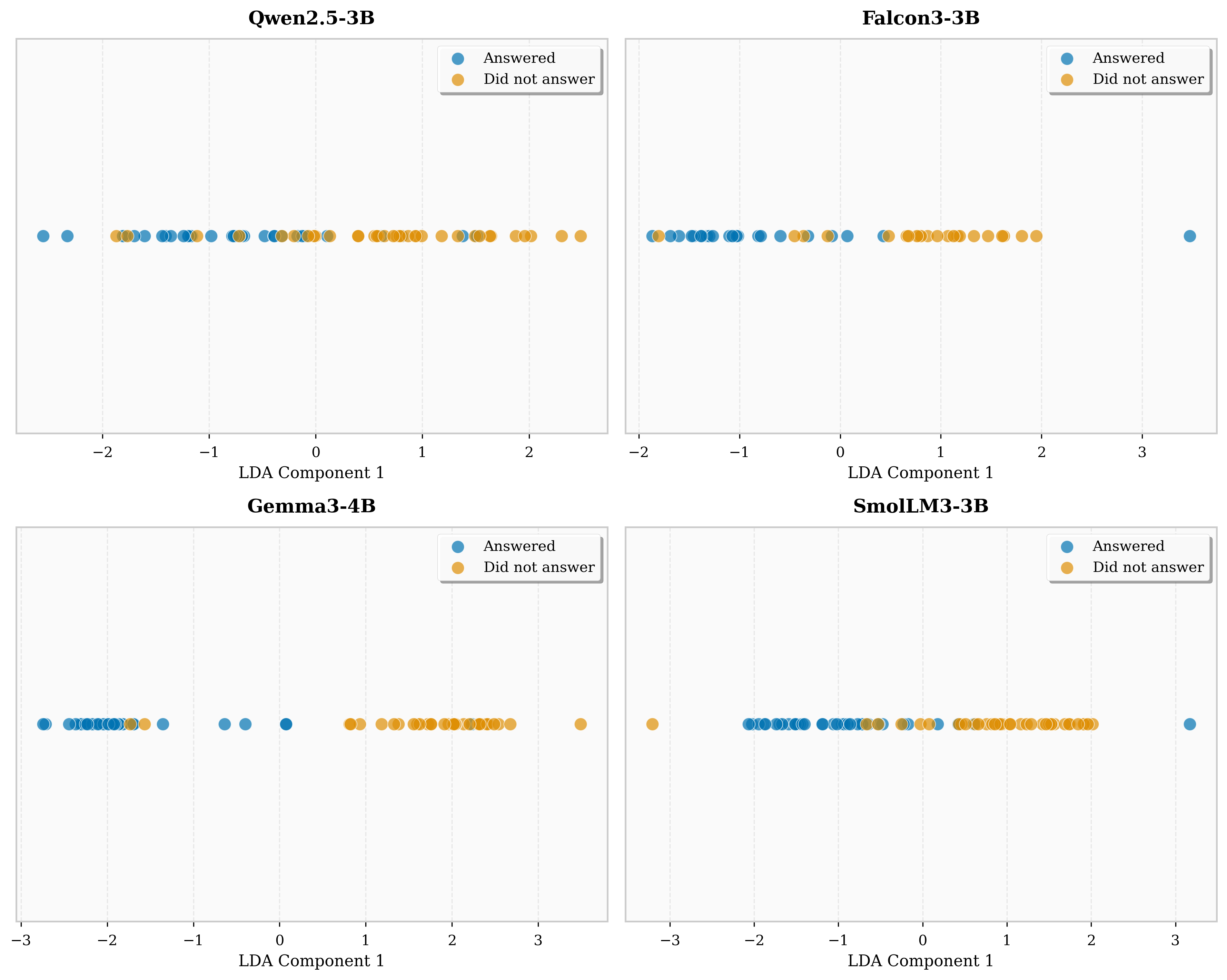}
    \caption{Ref: affirming, Train: affirming}
  \end{subfigure}
  \hfill
  \begin{subfigure}[b]{0.48\textwidth}
    \centering
    \includegraphics[width=\linewidth]{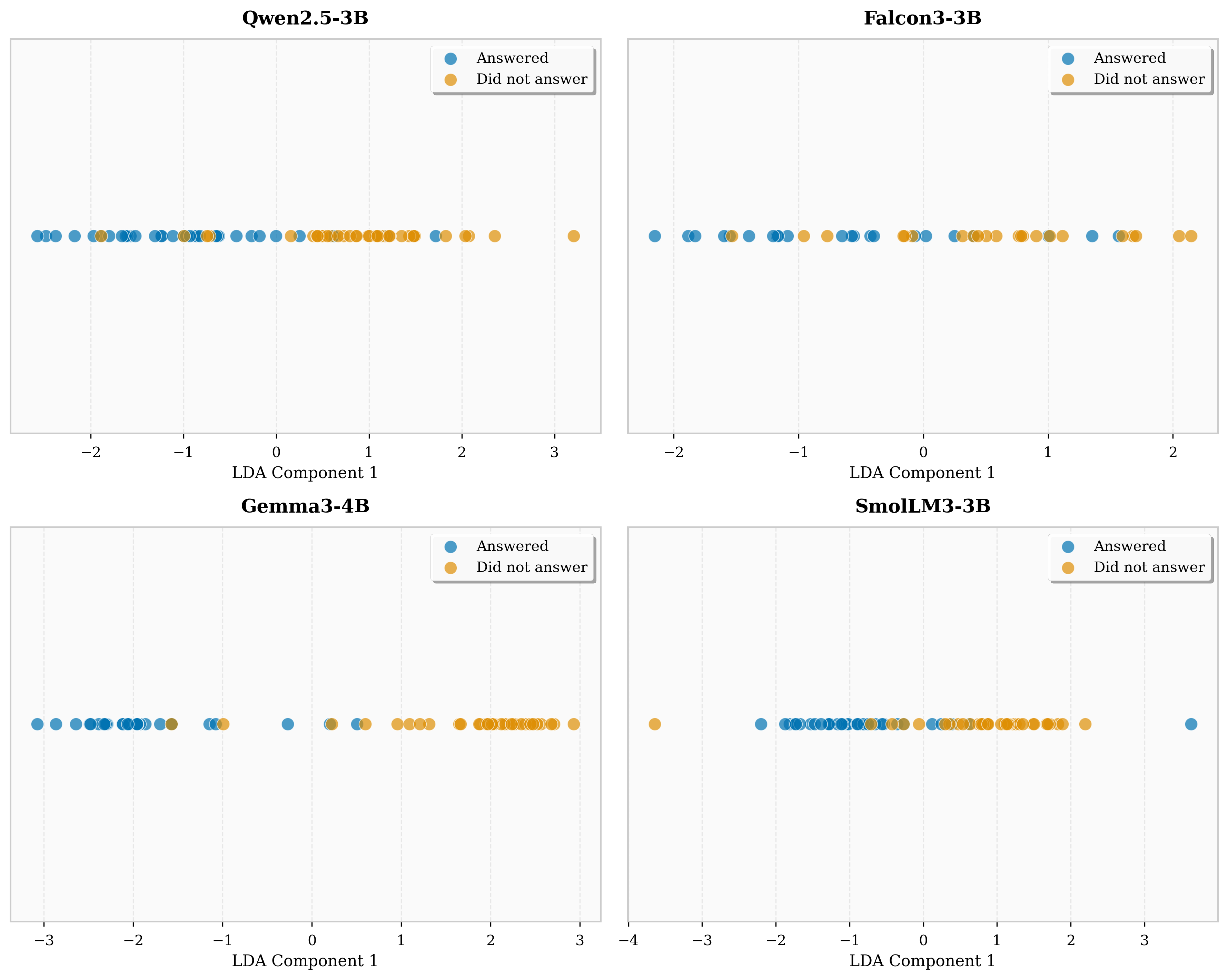}
    \caption{Ref: affirming, Train: rejection}
  \end{subfigure}

  \medskip

  \begin{subfigure}[b]{0.48\textwidth}
    \centering
    \includegraphics[width=\linewidth]{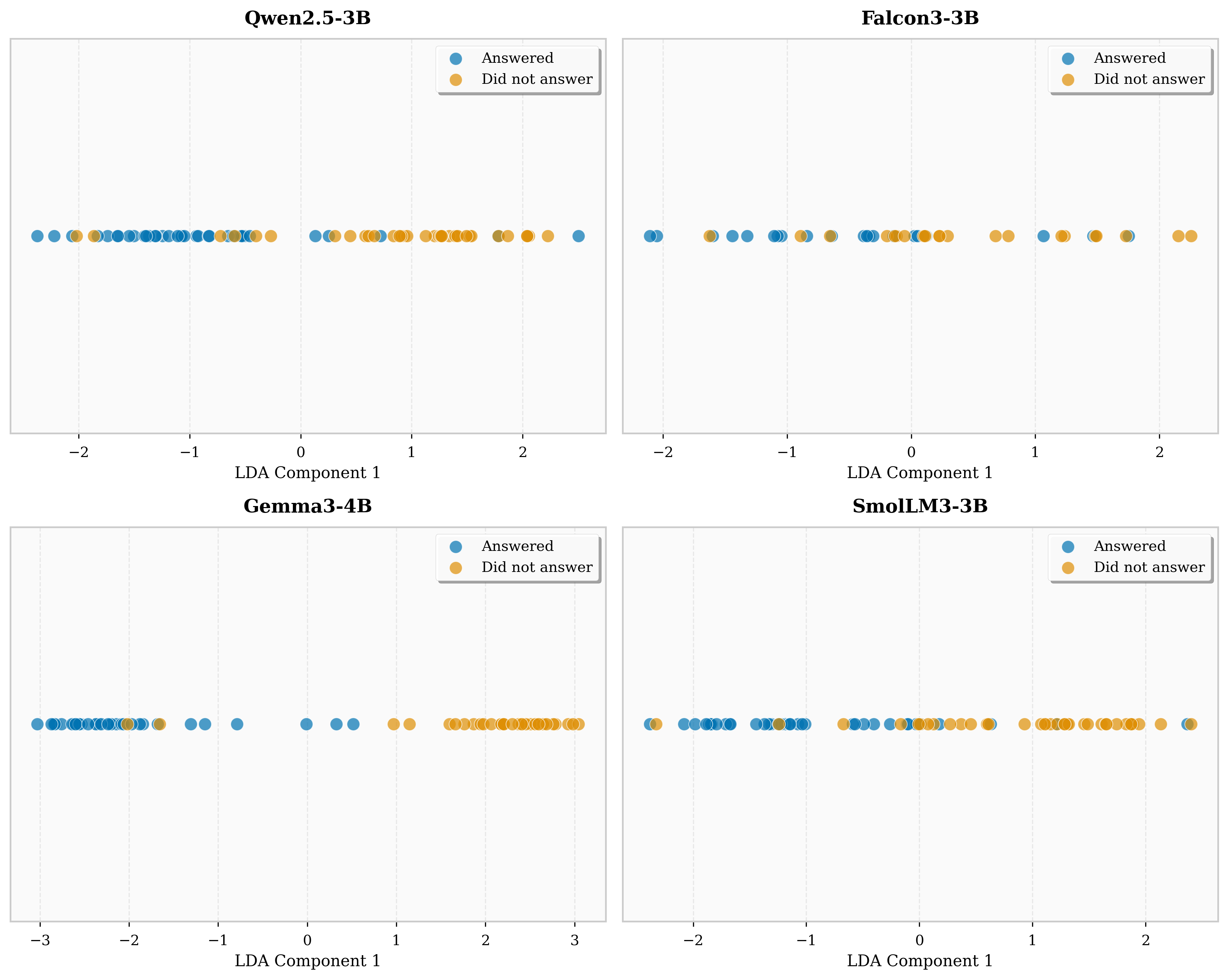}
    \caption{Ref: rejection, Train: affirming}
  \end{subfigure}
  \hfill
  \begin{subfigure}[b]{0.48\textwidth}
    \centering
    \includegraphics[width=\linewidth]{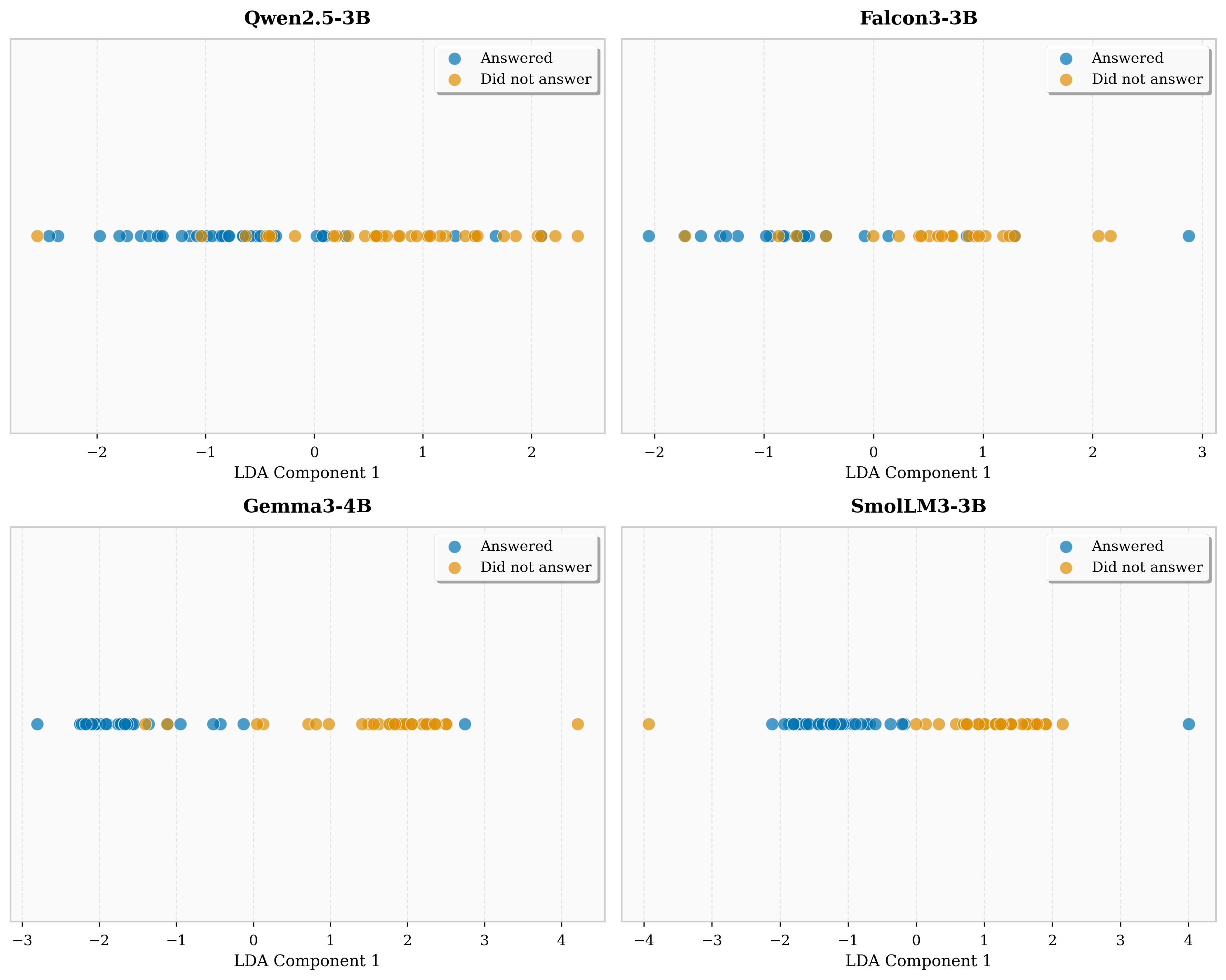}
    \caption{Ref: rejection, Train: rejection}
  \end{subfigure}
  \caption{Response task --- \textbf{Did Not Answer} reference
  gradient. LDA projections of layer-wise cosine similarities.
  The DNA reference produces the widest separation, confirming
  that abstention is the most distinctive behavioral mode in
  gradient space.}
  \label{fig:lda_resp_dna}
\end{figure*}

Figures~\ref{fig:lda_resp_correct} and~\ref{fig:lda_resp_dna}
show LDA projections for the binary Response task. Consistent
with the near-perfect classification accuracy reported in
Table~\ref{tab:response_results}, the Answered and Did Not Answer
clusters are widely separated across all reference classes and
response-type combinations. The separation is notably larger than
that observed in the Correctness task
(Figures~\ref{fig:lda_corr_correct}
and~\ref{fig:lda_corr_incorrect}), visually confirming that
abstention produces a qualitatively distinct gradient signature.

\subsection{Summary of Visualization Observations}
\label{app:lda_summary}

Several patterns are consistent across all visualizations.

\paragraph{Abstention is always well-separated.}
In every task and response-type combination, the Did Not Answer
cluster occupies a distinct region of the LDA projection. This
visual separation corresponds directly to the 94--99\% accuracy
observed on the Response task
(Table~\ref{tab:response_results}).

\paragraph{Correct vs.\ Incorrect separation is tighter.}
The Correct and Incorrect clusters overlap more than either does
with the DNA cluster, consistent with our finding that the
Correctness distinction is the primary bottleneck in three-way
classification. The overlap is greatest for the rejection
response type, where the model's abstention behavior reduces the
diversity of gradient patterns available for distinguishing
correctness.

\paragraph{Affirming references produce cleaner projections.}
Plots using affirming reference samples (subfigures (a) and (b)
in each figure) generally show tighter, better-separated
clusters than those using rejection samples. A single
configuration (e.g., \emph{Correct} reference with
affirming/affirming probes) is sufficient for deployment, as
predictors trained on each of the twelve configurations achieve
accuracy within a 2-point range.

\paragraph{Patterns are consistent across model families.}
All four models show qualitatively similar cluster structures
within each plot, confirming the cross-family consistency of
gradient-based behavioral signatures reported in
Section~\ref{sec:overall}.

\paragraph{All twelve configurations yield comparable separation.}
Across the $|\mathcal{C}| \times 2 \times 2 = 12$ combinations of
reference category, reference probe response, and training probe
response, the LDA projections show qualitatively similar cluster
structures.
Predictors trained on each of the twelve configurations achieve
accuracy within a 2-point range, confirming that the discriminative
gradient signal is robust to the choice of probe combination.
This invariance simplifies deployment: a single configuration
(e.g., \emph{Correct} reference, affirming/affirming probes) is